\documentclass{article} % For LaTeX2e
\usepackage{iclr2021_conference,times}

\usepackage{amsmath,amsfonts,bm}

\def\eqref#1{equation~\ref{#1}}
\def\1{\bm{1}}

\DeclareMathAlphabet{\mathsfit}{\encodingdefault}{\sfdefault}{m}{sl}
\SetMathAlphabet{\mathsfit}{bold}{\encodingdefault}{\sfdefault}{bx}{n}

\usepackage[utf8]{inputenc} % allow utf-8 input
\usepackage[T1]{fontenc}    % use 8-bit T1 fonts
\usepackage{hyperref}
\usepackage{url}
\usepackage{booktabs}       % professional-quality tables
\usepackage{amsfonts}       % blackboard math symbols
\usepackage{nicefrac}       % compact symbols for 1/2, etc.
\usepackage{microtype}      % microtypography
\usepackage{wrapfig,lipsum}
\usepackage{enumitem}
\usepackage{pifont}
\newcommand{\cmark}{\ding{51}}%
\newcommand{\xmark}{\ding{55}}%
\usepackage{textcomp}
\usepackage{subfigure}
\usepackage{amsmath,amssymb}
\usepackage{varwidth}
\usepackage{booktabs}
\usepackage{arydshln}
\usepackage{graphicx}
\usepackage{multirow}
\usepackage[T1]{fontenc}

\usepackage{bm}
\usepackage[font=footnotesize]{caption}

\usepackage{amssymb} 

\usepackage[utf8]{inputenc} % allow utf-8 input
\usepackage[T1]{fontenc}    % use 8-bit T1 fonts
\usepackage{hyperref}       % hyperlinks
\usepackage{booktabs}       % professional-quality tables
\usepackage{amsfonts}       % blackboard math symbols
\usepackage{nicefrac}       % compact symbols for 1/2, etc.
\usepackage{microtype}      % microtypography
\usepackage{wrapfig,lipsum}
\usepackage{import}
\usepackage{hhline}

\newcommand\nl[1]{\emph{``#1''}}

\newif\ifcomments
\commentstrue 
\ifcomments
    \providecommand\at[1]{[\textcolor{blue}{{AT: #1}}]}
    \providecommand\jb[1]{[\textcolor{red}{{JB: #1}}]}
    \providecommand\as[1]{[\textcolor{magenta}{{AA: #1}}]}
    \providecommand\oy[1]{[\textcolor{purple}{{OY: #1}}]}
    \providecommand\hh[1]{[\textcolor{olive}{{HH: #1}}]}
    \providecommand\ac[1]{[\textcolor{brown}{{AC: #1}}]}
    \providecommand\dl[1]{[\textcolor{violet}{{DL: #1}}]}
    \providecommand\yw[1]{[\textcolor{orange}{{YW: #1}}]}
    \providecommand\gi[1]{[\textcolor{teal}{{GI: #1}}]}
\else
    \providecommand\at[1]{}
    \providecommand\jb[1]{}
    \providecommand\as[1]{}
    \providecommand\ac[1]{}
    \providecommand\hh[1]{}
    \providecommand\oy[1]{}
    \providecommand\dl[1]{}
    \providecommand\yw[1]{}
    \providecommand\gi[1]{}
\fi
\newcommand\comment[1]{}
\newcommand\bfemph[1]{\textbf{\emph{#1}}}

\newcommand{\WikiEnt}{\emph{WikiEntity}}
\newcommand{\WikiEnts}{\emph{WikiEntities}}
\newcommand{\MMQA}{\emph{MMQA}}
\newcommand{\ImplicitDecomp}{\emph{ImplicitDecomp}}
\newcommand{\AutoRouting}{\emph{AutoRouting}}
\newcommand{\ImageListQ}{ImageListQ}
\newcommand{\ImageQ}{ImageQ}
\newcommand{\TableQ}{TableQ}
\newcommand{\TextQ}{TextQ}

\title{MultiModalQA: complex question answering over text, tables and images}

\author{Alon Talmor\thanks{\;\;\; The authors contributed equally}$^{*,1,2}$ ~~ Ori Yoran$^{*,1,2}$ ~~
Amnon Catav$^{*,2}$ ~~
Dan Lahav$^{*,2}$ ~~
Yizhong Wang$^{3}$ \\
\textbf{Akari Asai}$^{3}$  ~~
\textbf{Gabriel Ilharco}$^{3}$ ~~
\textbf{Hannaneh Hajishirzi}$^{2,3}$ ~~
\textbf{Jonathan Berant}$^{1,2}$ \\
$^1$The Allen Institute for AI, ~$^2$Tel-Aviv University, ~$^3$University of Washington \\
\texttt{\{alont,oriy,jonathan\}@allenai.org} \\
\texttt{\{amnoncatav,lahav\}@mail.tau.ac.il}\\
\texttt{\{yizhongw,akari,gamaga,hannaneh\}@cs.washington.edu}
}

\iclrfinalcopy % Uncomment for camera-ready version, but NOT for submission.
\begin{document}
\maketitle
\begin{abstract}

%When answering a question, people often combine knowledge from visual, textual and tabular information. Recent work has focused on question answering that involves integrating multiple pieces of evidence, but did not require  simultaneous reasoning across multiple modalities. 
%In this paper, we propose a novel framework for generating complex multi-modal questions at scale.       
%To facilitate complex multi-modal question answering, 
%We present \textsc{MultiModalQA} (MMQA): a challenging new dataset that requires joint reasoning over text, tables and images. 
%The questions \& answers in MMQA are associated to contexts from Wikipedia, but can also be used in an open domain setup. 
%Each context contains an anchor table from Wikipedia, to which we connect images and relevant text questions from existing Reading Comprehension (RC) datasets. 
%Using the Wikipedia contexts, we automatically generate pseudo language multi-modal questions, and then used AMT workers to paraphrase the machine-generated questions.
%We create 26,878 questions through this procedure, and empirically demonstrate the necessity of a multi-modal approach to solve our task via \ImplicitDecomp{} a novel mutli-step model that achieves an average F1 of XXX over compositional questions that combine multiple modalities, significantly outperforming a single-step model that achieves XXX F1. 
%We hope our work will help attract new research on question answering involving multiple modalities. 

When answering complex questions, people can seamlessly combine information from visual, textual and tabular sources. 
While interest in models that reason over multiple pieces of evidence has surged in recent years, there has been relatively little work on question answering models that reason \emph{across} multiple modalities.
In this paper, we present \textsc{MultiModalQA} (MMQA): a challenging question answering dataset that requires joint reasoning over text, tables and images. 
We create \textsc{MMQA} using a new framework for generating complex multi-modal questions at scale,    
%To facilitate complex multi-modal question answering, 
harvesting tables from Wikipedia, and attaching images and text paragraphs using entities that appear in each table. We then define a formal language that allows us to take questions that can be answered from a single modality, and combine them to generate \emph{cross-modal} questions. Last, crowdsourcing workers take these automatically generated questions and rephrase them into more fluent language.
%Each context contains an anchor table from Wikipedia, to which we connect images and relevant text questions from existing Reading Comprehension (RC) datasets. 
%Using the Wikipedia contexts, we automatically generate pseudo language multi-modal questions, and then used AMT workers to paraphrase the machine-generated questions.
We create 29,918 questions through this procedure, and empirically demonstrate the necessity of a multi-modal multi-hop approach to solve our task:
our multi-hop model, 
\ImplicitDecomp{}, achieves an average F$_1$ of 51.7 over cross-modal questions, substantially outperforming a strong baseline that achieves 38.2 F$_1$, but still lags significantly behind human performance, which is at 90.1 F$_1$.

\end{abstract}
\section{Introduction}

When presented with complex questions, people often do not know in advance what source(s) of information are relevant for answering it. In general scenarios, these sources can encompass multiple modalities, be it paragraphs of text, structured tables, images or combinations of those. For instance, a user might ponder
\nl{When was the famous painting with two touching fingers completed?},
%\nl{When will the film, showing half a lady’s face on it’s poster, be in cinemas?},
%\nl{What was the release year of Ben Piazza's movie with half a lady's face on the poster?},
if she cannot remember the exact name of the painting. Answering this question is made possible by integrating information across both the textual and visual modalities.

%JB: tried to make it a bit more captivating.
% When people look for an answer to a complex question, they often do not  know in advance what is the relevant source of information, be it a paragraph of text, a structured table, or an image. For instance, a user might look for an answer to the question \emph{What was the release year of Ben Piazza's movie with half a lady's face on the poster?}, if she cannot remember the exact name of the movie. Such questions require integrating information across both the textual and visual modalities.

%Humans often want to answer complex questions that require integrating information from several modalities, i.e. modes of expression. For instance, the question \emph{What was the release year of Ben Piazza's movie with half a lady's face on the poster?} requires one to search visual and textual information and to reason over both.

Recently, there has been substantial interest in question answering (QA) models that reason over multiple pieces of evidence (\emph{multi-hop} questions \citep{yang2018hotpotqa,talmor2018web,welbl2017constructing}). In most prior work, the question is phrased in natural language and the answer is in a context, which may be a paragraph \citep{squad2016url}, a table \citep{pasupat2015compositional}, or an image \citep{antol2015vqa}. However, there has been relatively little work on answering questions that require integrating information \emph{across} modalities.
\citet{hannan2020manymodalqa} created \textsc{ManyModalQA}: a dataset where the context for each question includes information from multiple modalities. However, the answer to each question can be derived from a {\it single modality} only, and no \emph{cross-modality reasoning} is needed. Thus, the task is focused on identifying the relevant modality.
%where an agent must answer a question by considering distinct modalities. However, in their work the solver is not required to answer questions that demand reasoning across modalties, rather the dataset is composed of single-modality questions, and the task is strictly understanding the modality of the question.
Recently, \citet{chen2020hybridqa} presented \textsc{HybridQA}, a dataset that requires reasoning over tabular and textual data. While \textsc{HybridQA} requires cross-modal reasoning, it does not require visual inference,  limiting the types of questions that can be represented (See Table~\ref{tab:datasets} for a comparison between the datasets).

%Thus, such efforts highlight the need for multimodal datasets and models that will facilitate reasoning over them.     \at{i think this paragraph can be shorter}

%exploring the scenario where the context includes multiple modalities and reasoning must occur across modalities.

%DL: original longer version
%Earlier work in question answering (QA) across modalities resorted to constructing datasets of single-modality questions in which the challenge is understanding the modality of the question \cite{hannan2020manymodalqa}. \at{many modal qa is just one example, "In most works" is already used avoce, perhaps shorten this peragraph and just say \cite{hannan2020manymodalqa} did this ... however ... } However, in such work the solver is not required to answer complex questions that demand reasoning across modalties, rather the task is strictly clearing the ambiguity in order to figure the modality of origin.  \at{i think many modal is just one example}
%Lately, \citet{chen2020hybridqa} proposed HybridQA, a dataset that requires reasoning over tabular and textual data. While HybridQA necessitates multimodal reasoning, it does not require visual inference which significantly limits the scope of the types of questions that can be represented in the dataset. Thus, such efforts highlight the need for multimodal datasets and models that will facilitate reasoning over them.     \at{i think this paragraph can be shorter}

\begin{figure}[t]
  \includegraphics[width=1.0\columnwidth]{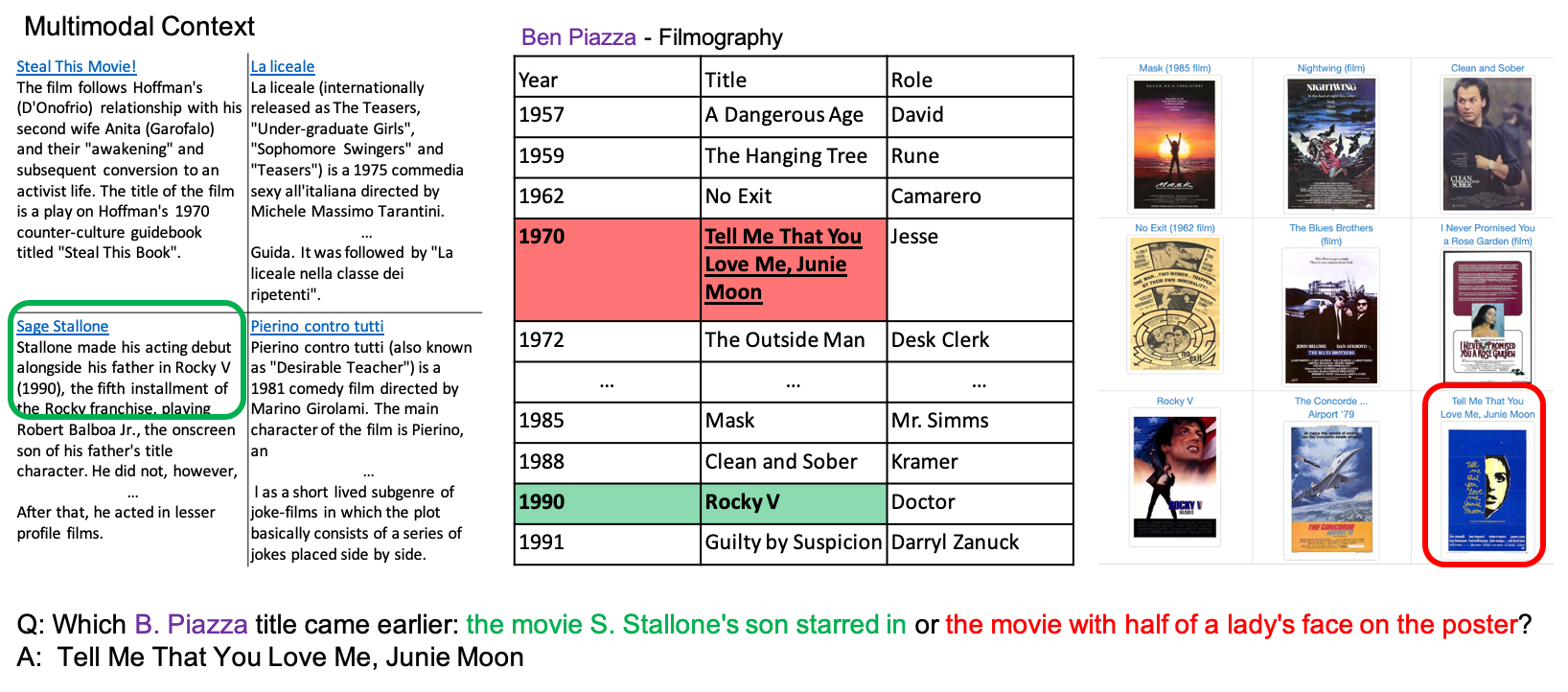}
  \caption{Example of a \MMQA{} question, answer and context. In green are the text modality question and answer, and in red the image modality. The table is used to perform the year comparison between the answers of the text and image question parts. }
  ~\label{fig:intro-fig}
  \vspace{-10pt}
\end{figure}

In this work, we present \MMQA{}, the first large-scale (29,918 examples) QA dataset that requires integrating information across free text, semi-structured tables, and images, where 35.7\% of the questions require cross-modality reasoning.  
Figure \ref{fig:intro-fig} shows an example question: \nl{Which B.Piazza title came earlier: the movie S. Stallon's son starred in, or the movie with half of a lady's face on the poster?}.
Answering this question entails
(i) decomposing the question into a sequence of simpler questions, (ii) determining the modalities for the simpler questions and answering them, i.e., information on the poster is in an image, the information on \nl{S. Stallon's son} is in free text, and the years of the movies are in the table, (iii) combining the information from the simpler questions to compute the answer: \nl{Tell Me that you love me, Junie Moon}.

Our methodology for creating \MMQA{} involves three high-level steps. (a) \emph{Context construction}: we harvest tables from Wikipedia, and connect each table to images and paragraphs that appear in existing Reading Comprehension (RC) datasets \citep{kwiatkowski2019natural, clark2019boolq, yang2018hotpotqa}; (b) \emph{Question generation}: Following past work \citep{talmor2018web}, we use the linked structure of the context to automatically generate questions that require multiple reasoning operations (composition, conjunction, comparison) across modalities in pseudo-language ; (c) \emph{Paraphrasing}: we use crowdsourcing workers to paraphrase 
%\citep{wang2015overnight} 
the pseudo-language questions into more fluent English.

\begin{wraptable}[13]{R}{0.5\columnwidth}
\vspace{-5pt}
\begin{center}
\resizebox{0.5\columnwidth}{!}{
\begin{tabular}{l|c|c|c|c}
Dataset     & Size  & Full- & Uses & Multi- \\ 
            &   & wiki & images & hop \\ 
\hline & & & \\[-1.7ex]
\textsc{ManyModalQA} &  10K & \xmark & \cmark & \xmark \\ 
& & & \\[-1.7ex]
\textsc{HybridQA}    &  70K & \xmark & 
\xmark & \cmark \\ 
\hline & & & \\[-1.7ex]
\textsc{MultiModalQA}&  30K & \cmark & \cmark & \cmark \\ 
\end{tabular}}
\end{center}
\caption{A comparison of \textsc{MultiModalQA} to 
\textsc{ManyModalQA} and \textsc{HybridQA}. We compare dataset size, use of images, and whether the dataset supports multi-hop questions and an open-domain full-wiki setup.}
\label{tab:datasets}
\vspace{5pt}
\end{wraptable}

%Following \cite{talmor2018web}, propose a framework for generating complex multimodal questions at scale. We build a dataset of questions \& answers that are associated to contexts from Wikipedia, but can also be used in an open domain setup (\cite{chen2017reading}). Each context contains an anchor table from Wikipedia, to which we connect images via the Wikipedia links of the table and relevant text questions from existing Reading Comprehension (RC) datasets (\cite{kwiatkowski2019natural, clark2019boolq, yang2018hotpotqa}). Using the context, we automatically generate pseudo language multimodal questions that include reasoning types as composition, conjunction and comparative. We then use AMT workers to paraphrase the machine-generated questions into natural spoken language. 

To tackle \MMQA{}, we introduce \ImplicitDecomp{}, a new model that predicts a program that specifies the required reasoning steps over different modalities, and executes the program with dedicated text, table, and image models. \ImplicitDecomp{} performs multi-hop multimodal reasoning without the need for an explicit decomposition of the question.

%We experiment with several baseline models ranging from using only one modality to infer the answer, to more complex architectures that access information scattered across modalities. We introduce \ImplicitDecomp{}, a new model that is capable of reasoning over multiple modalities by laying out a program for compositional operations over modalities, and executing using text, table and image models in a multi-hop fashion, without the need for explicit decomposition of the question.  
We empirically evaluate \MMQA{} by comparing \ImplicitDecomp{} to strong baselines that do not perform cross-modal reasoning and to human performance. We find that on multimodal questions, \ImplicitDecomp{} improves F$_1$ from $38.2  \rightarrow 51.7$ over a single-hop approach. Humans are able to reach 90.1 F$_1$, significantly outperforming our best model. Because automatic evaluation is non-trivial, we also manually analyze human performance and find humans correctly answer 94.5\% of the questions in \MMQA{}.
Finally, our dataset can be used in an open-domain setup over all of  Wikipedia. In this setup, the F$_1$ of humans is 84.8.

%We evaluate our dataset using the baseline models and an additional expert-human baseline. We find that on multimodal questions, \ImplicitDecomp{} improve performance from 42.9 to 51.1 over a simple single modality approach. Humans are able to reach XXX performance, significantly outperforming our best baseline. Finally, our dataset can be used over the whole Wikipedia in an open domain setup, for which humans achieve an accuracy of XXX.
%\gi{add some comment on making it publicly available?}  
\pagebreak
To summarize, our key contributions are:
\begin{itemize}[topsep=0pt,noitemsep,leftmargin=4mm]
    \item \MMQA{}: a dataset with 29,918 questions and answers, 35.7\% of which require cross-modal reasoning.%  that will be made publicly available. 
    \item A methodology for generating multimodal questions over text, tables and images at scale.
    \item \ImplicitDecomp{}, A model for implicitly decomposing multimodal questions, which improves on a single-hop model by 13.5 absolute F$_1$ points on questions requiring cross-modal reasoning.
    %\jb{I hink single modality here sounds like it is using just one modality, but autorouting is using all of them, so you are underselling here. Maybe just say 'on a strong baseline'}. \at{perhaps \nl{simple single hop} ? }
    \item Our dataset and code are available at \url{https://allenai.github.io/multimodalqa}.
\end{itemize}

\section{Dataset Generation}

\begin{figure}[h]
  \includegraphics[width=1.0\columnwidth]{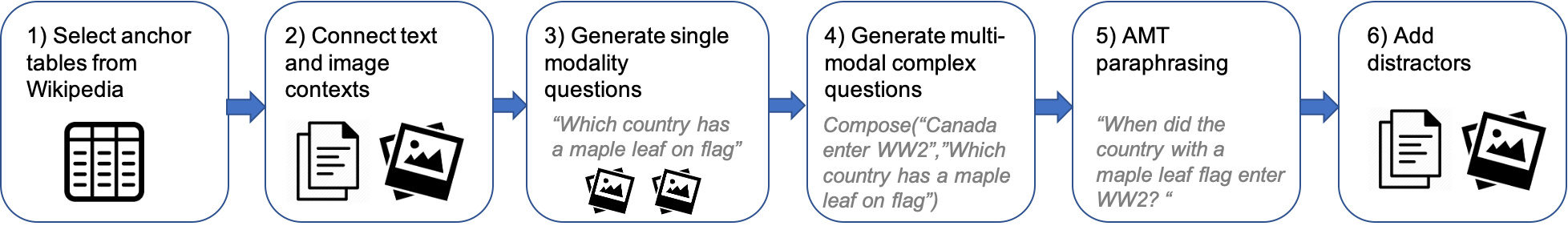}
  \caption{An overview of \MMQA{} dataset generation process.} 
  %\gi{Nit: the arrows between the boxes are not aligned properly (some are overlapping with the boxes)} \jb{why caps for 'Paraphrasing' and 'Distractors'?}}
  ~\label{fig:dataset-generation}
  \vspace{-10pt}
\end{figure}

Our goal is to develop a method that allows generating complex questions over multiple modalities at scale. An overview of the methodology is captured in Figure \ref{fig:dataset-generation}. 
We first select a Wikipedia table as an anchor, to which we add images and texts paragraphs and obtain a \emph{context}. Single modality questions are generated based on these contexts, and used to automatically create multimodal, multi-hop questions. AMT workers rephrase the questions into natural language, and finally distractor paragraphs and images are selected for each question.
We now elaborate on the $6$ steps of the process.

%We generate the questions via the following framework:
%\begin{enumerate}%[topsep=0pt,itemsep=0ex,parsep=0ex]
%    \item We extract tables from Wikipedia, and use these tables as the anchors of the context.
%    \item We connect images to the context via the Wikipedia links of the table, and ask AMT workers to author questions on the images.
%    \item We connect relevant text questions, to the context, from existing RC datasets, via the Wikipedia links of the table.
%    \item Using the context, we automatically generate pseudo language multimodal questions that include reasoning types as composition, conjunction and comparative.
%    \item We use AMT workers to paraphrase the machine-generated questions, and we add distractors to each question.
%\end{enumerate}
%\at{i think given the figure this is a bit redundant. Can we just say some thing like \nl{Figure 2 describes the framework we used to create the dataset. The remaining parts of the section elaborates on each of the five steps. }} \dl{do you mean to get rid of the enumeration above?}

%To that end, we build a dataset that consists of questions\&answers that are associated to contexts. 
%Each context contains: (i) a single table; (ii) images related to the table; (iii) text paragraphs related to the questions (both gold and distractors). \dl{I think the above is redundant given that we added a description of the process. if you agree please cut.}

%\at{TODO describe in general the process and refer to figure}

%\jb{please number paragraphs with latex and labels for backreference? \url{https://tex.stackexchange.com/questions/129208/numbering-paragraphs-in-latex}}
\textbf{2.1 Wikipedia tables as anchors}
%\at{i think we can move to paragraphs with numbers corresponding to boxes in fig 2 (that will also be numbered) instead of most subsections, to save the white spaces...}
The 01-01-2020 English Wikipedia dump contains roughly $3M$ tables. We extracted all tables and selected those that meet the following criteria: (a) The tables contain 10-25 rows
%, i.e. they are not too short or too long; 
(b) At least 3 images are associated with the table. This results in a total of 700k tables.
(see supp. material for more information). 
These tables are the anchors of our contexts, which we enrich with images and text for multimodal question generation. 
A key element of the tables are Wikipedia Entities (\WikiEnts{}) that appear in them, i.e., concepts linked to other Wikipedia entries. We 
%denote \WikiEnts{} as \WikiEnt{}, and 
use them to connect different modalities, bridge questions, and solve ambiguities (details below).
%\dl{last sentence may need refining}
%\jb{I am not convinced having this WikiEnt definition is needed. it makes the text a bit cumbersome where I think we can just talk about \WikiEnts{} where needed.}

%\dl{for space reasons we removed the classifier section; we need to remember to address it briefly when discussing images and MG}
%On top of filtering, we use heuristics to determine the \emph{semantic type} of each table column: \jb{say what are the types here if short or in footnote?}. Types are used in \S \jb{where? ref to subsection} when adding elements to the tables \jb{why do we add elements to the tables?}, and when constructing questions. \dl{AT, since we are lacking space and don't really go into detail about the classifier, upon reflection maybe it is best to just move this paragraph to the supp?}

%\subsection{Adding the Visual Modality}
\textbf{2.2 Connecting Images and Text to Tables} 
%\gi{perhaps a high-level sentence about this step so the transition is less dry?}
\bfemph{Images.} We consider two cases: (a) in-table images and (b) images from pages of linked \WikiEnts{}. 
In the former, the images are featured inside the table cells.
%, thus we classify the column as containing images and retrieve the images.
In the latter, the table contains a column of \WikiEnts{} that potentially have images, e.g. a table describing the filmography of an actor often contains a 
%\WikiEnt{} \jb{do we need WikiEnt here?} 
column of film names, which may have posters in their respective pages. To associate entities with their representative image, we map entities and their profile images in their Wikipedia pages. Overall, we obtain 57,713 images, with 889 in-table images and 56,824 \WikiEnts{} images. 
%Due to a variety of reasons, as missing images in Wikipedia, this process could not be done for all \WikiEnt{}. \dl{I think the last sentence may go to supp} \at{yep, we can add "to most tables" in some sentence earlier}. 
%DL: changed it to broadly and we can have a ref to supplementary if need be 
\bfemph{Text.} We build on texts from contexts appearing in existing reading comprehension datasets. We elaborate on this process next.

\textbf{2.3 Generating Single-Modality Questions}
%\gi{similarly to the previous section, I think a high-level sentence would help readability here}
\bfemph{Tables.} We generate pseudo-language table questions in the following form \nl{In [table title] of [Wikipedia page title] which cells in [column X] have the [value Y] in [column Z]?}. We additionally support numeric computations over columns classified as dates or numbers, such as min and max values, e.g., \nl{In [Doubles] of [WCT Tournament of Champions], what was the MOST RECENT [Year](s) where the [Location] was [Forest Hills]}. 

\bfemph{Images.} We use crowdsourcing to generate single-modality questions about images. We generated two types of image questions, based on the images we retrieved from the previous step: (i) questions over a single image, (ii) questions over a list of images. 

When generating single-image questions, we show Amazon Mechanical Turk (AMT) crowd workers an image alongside its \WikiEnt{}, and ask them to phrase a question about the image with the entity being the focus of the question. E.g, if the entity is \nl{Roger Federer}, a potential question is \nl{What's the hair color of Roger Federer?}. For questions to have meaning in an open-domain setting, we primed AMT workers to ask questions that correspond to ``stable'' features, i.e., features that are unlikely to change in different images and are thus appropriate in an open-domain setting. 
%\jb{it seems these questions are single modality. Is this true? if so, say it explicitly}
%(e.g. the previous question rather than \nl{What is Roger Federer holding?}). 

For questions with a list of images, we use images that appear in the same column of a table. To generate these questions, AMT workers were given the images and asked to phrase a binary question about a distinctive feature of the entities 
%\WikiEnt{} \jb{entities?} 
that a subset of the images share. E.g., given a list of statues, the worker could ask \nl{Which of the statues features a horse?} 
%\jb{this does not sound open-domain since an image might just not how a garden though it exists}
This process results in 2,764 single image questions and 7,773 list image questions that are later used to create multimodal questions.

\bfemph{Text.} To obtain questions answerable over text paragraphs we build on existing reading comprehension datasets: \emph{Natural Questions (NQ)} \citep{kwiatkowski2019natural} consists of about $300K$ questions issued to the Google search engine. 
%The questions in this dataset were authored by real anonymized users and represent the natural distribution. 
This dataset mostly contains simple questions where a single paragraph suffices to answer each question. \emph{BoolQ} \citep{clark2019boolq} contains $15,942$ yes/no questions, gathered using the same pipeline as NQ. \emph{HotpotQA} \citep{yang2018hotpotqa} contains $112K$ training questions, where crowd workers were shown pairs of related Wikipedia paragraphs and were asked to author questions that require multi-hop reasoning over the paragraphs.
%\jb{is there another version?}
%We use the questions, answers, and gold paragraphs of the distractors version of this dataset which contains about 112K questions. 
%\dl{'the distractors version' is a little cluttered, we should probably define or remove it} \ac{I think we can make this paragraph shorter}

To use questions from the above datasets as building blocks for multi-hop multimodal questions, we unified them into a corpus that consists of triples of (i) a text question, (ii) an answer and (iii) 1-2 gold paragraphs from Wikipedia. We link a question to a table, by matching \WikiEnts{} in the table to entities in the text of the question (see supplementary material for further details). 
%(the entities are tagged by Wikipedia), 
%\jb{didn't understand this - the question is not part of wikipedia how can it be tagged by it}) \at{let's say the entities in table are marked by wikipedia, and ya how do mark the questions in the NQ questions? Dan can you ask amnon about this?}. 
%\jb{not using any tool for entity detection?} 
%\dl{the process is as follows: (1) we take the html of the gold article; (2) we extract all the Wiki entities from the html (they are tagged); (3) we search for a match between the tokens of each entity and the questions (using spacy); (4) if a match is found, we mark the entity as one that appears in the question; (5) if an entity is shared between the question and table we connect them (this step is deterministic because we use the unique title);}
%We use a context aware string matching heuristic approach to match entities between the tables and questions (see supplementary material for further details).
Overall, we retrieved 6,644 questions from NQ, 1,246 from BoolQ and 4,733 from HotpotQA.%\ac{I'm not sure if the word "retrieved" here is clear enough. This is the amount of questions we eventually used in the dataset, not the amount we retrieved from the external sources/linked at the first step to the tables. Maybe replace with "leveraged"?}

\textbf{2.4 Generating multimodal complex questions}
We present an automatic method for creating at scale multimodal compositional questions (i,e., questions that require answering a sequence of sub-questions to conclude the final answer). Our first step is to introduce a formal language that allows to combine questions answerable from a single modality. 
Below we introduce the logical operations that allow to generate such pseudo-language (PL) questions, while keeping a formal representation of how they were constructed. 
In Table~\ref{tab:compositionality-breakdown}, we illustrate this process with all 16 different compositional templates used for question generation. We now describe our logical operations.

%\at{ 2. It's probably helpful to have a figure showing the create of on complex PL question... }
\begin{table*}[t]
\centering
\resizebox{1.0\textwidth}{!}{
\begin{tabular}{l|l|c}
\hline
\textbf{Type}                     & \textbf{Q\&A}                                                                                                                                                                                                                     & \textbf{\%} \\ \hline
%\TextQ{}  \textit{(simple)}            & \textit{who is the no 1 artist on spotify? Drake}     & 15.3    \\ \hline
% \textit{(complex)}
\TextQ{}        & \textit{What was the territorial capital of the territory opposing Ohio} & 31.0    \\ 
& \textit{in the Toledo War? Detroit} & \\ \hline

% \textit{(complex)}
\TableQ{}   & \textit{Does the German state Baden-Wurttemberg or Thuringia have more} & 18.3      \\
& \textit{  residents? Baden-Württemberg} & \\ \hline

\ImageQ{}  & \textit{What weapon is the statue in Nottingham holding? bow} & 8.9      \\ \hline

Compose(\TextQ{},\TableQ{})  & \textit{At what age did the Cleveland Cavaliers player with 6190} & 7.8     \\ 
& \textit{rebounds enter the NBA? 19} & \\ \hline

%\TableQ{}  & \textit{What club did Mesut Ozil play for in La Liga in 2010-11? Real Madrid} & 9.3      \\ \hline

\ImageListQ{} & \textit{What is the common name of the bush warbler in Thailand that has an}                                                                                   & 6.1     \\ 
& \textit{  orange stripe above its eye?  Chestnut-crowned bush warbler} & \\
\hline

%Compose(\TableQ{},\TableQ{}) & \textit{What category did The Coca-Cola Kid win an award for at the Cannes} & 4.8      \\
%& \textit{ Film Festival? Palme dOr} & \\ \hline

Compose(\TableQ{},\ImageListQ{}) & \textit{The film that starred Chris Ellison where a man was holding a} & 5.4 \\ 
& \textit{  newspaper on the poster, was released what year? 1988} & \\ \hline

%Intersect(\TableQ{},\TableQ{}) & \textit{How long is the Vazhvile Oru Naal Soundtrack song written by Ku. Sa.} & 4.1      \\ 
%& \textit{ Krishnamoorthy and sung by Jikki? 06:00} & \\ \hline

Compose(\ImageQ{},\TableQ{}) & \textit{On the poster for the TV show in which Tom Mison played Dorian}  & 4.5      \\
& \textit{Crane, what kind of structure can be seen behind the two men? castle } & \\ \hline

%Compare(\TableQ{},\TableQ{}) & \textit{Does the German state Baden-Wurttemberg or Thuringia have more} & 3.3      \\
%& \textit{  residents? Baden-Württemberg} & \\ \hline

Compare(Compose(\TableQ{},\ImageQ{}),\TableQ{})  & \textit{Which manufacturer has fewer wins at the First Data 500: Buick or the}                                                                                                                                                                               & 3.5      \\ 
& \textit{  brand with a cross for a logo? Buick} & \\
\hline

Compose(\TableQ{},\TextQ{})  & \textit{On what date did the original artist who sang Sweet Child of Mine have} & 3.2      \\
& \textit{  a concert at US Bank Stadium? July 30, 2017} & \\ \hline

Intersect(\TableQ{},\TextQ{})  & \textit{Who was the artist for Damon Fox in 2006 who also sings "You got the }                                                                                                      & 2.6      \\
& \textit{moves like Jagger"? Christina Aguilera} & \\
\hline

Compose(\TextQ{},\ImageListQ{})  & \textit{On the poster for the movie based on the book "Act like a Lady, Think Like}                                                                                       & 2.4      \\ 
& \textit{a Man," how many people are there in total? nine} & \\
\hline

Intersect(\ImageListQ{},\TableQ{}) & \textit{What covers of the Chandler Canterbury films from 2009 has  more than}                                                                 & 2.3      \\
& \textit{one person? Powder Blue  Balls Out,  Gary the Tennis Coach, After.Life} & \\
\hline

Compare(\TableQ{},Compose(\TableQ{},\TextQ{}))  & \textit{Did Chelsea or club that sings You'll Never Walk Alone rank higher in}                                                                                               & 2.1     \\
& \textit{Deloitte Football Money League 2007? Chelsea} & \\
\hline

Compose(\ImageQ{},\TextQ{})  & \textit{Did Gary Oldman take part in the movie whose poster features  two men}                                                                         & 1.0    \\
& \textit{ holding handguns, and which had Mark L. Smith as a writer? no} & \\
\hline

Compare(Compose(\TableQ{},\ImageQ{}),   & \textit{
Was the film that features a giant eye on its poster or the first Wolverine} & 0.8     \\
   Compose(\TableQ{},\TextQ{})) & \textit{ movie the earlier film that Scott Silver worked on? Requiem for a Dream} & \\ \hline

Intersect(\ImageListQ{},\TextQ{}) & \textit{
What common law state with an eagle on the flag has an institution} & 0.2 \\
& \textit{ in the North region of Division II of the NCCAA? Iowa} & \\ \hline

\end{tabular}}
\caption{All 16 compositional templates in \MMQA{} with an example and their relative frequency. 
%\jb{About TableQ(copmlex): when talking about single-modality questions this was not mentioned, so a bit weird here.}.
}
\label{tab:compositionality-breakdown}
\vspace{-10pt}
\end{table*}

\textbf{Logical Operations}
Functions in our formal language take arguments and return a PL partial question, as well as answers that can be a list of one or more strings, or a list of one or more \WikiEnts{}.
%(that can be converted to strings for human readability and evaluation)
All operations have access to the full context. In addition, we prepend a prefix containing the Wikipedia table name and page title---e.g. \nl{In the Filmography of Brad Pitt,}---to all our PL questions to support an open-domain QA setup. Our set of logical operations are:
\begin{enumerate}[nosep,leftmargin=4mm]
\item \textbf{\textsc{\TableQ{}}}:
Returns a question from the table questions generated in \S2.3, as well as a list of \WikiEnts{} or a list of strings as answers.% Additionally, we compose single modality table questions using the operations described below. 
\item \textbf{\textsc{\TextQ{}}}:
Returns a text corpus question (see \S2.3) and a list of % one or more
\WikiEnts{} or strings as answers.
\item \textbf{\textsc{\ImageQ{}}}:
Returns a question about a single image associated with a \WikiEnt{} and a single token answer from a fixed vocabulary (see \S2.3).
\item \textbf{\textsc{\ImageListQ{}}}:
Returns a question about a list of images 
%that are associated with a table column containing \WikiEnts{} 
and a list of \WikiEnts{} corresponding to the images that answer the question (see \S2.3).
\item \textbf{\textsc{Compose$(\cdot, \cdot)$}}:
Takes a PL question containing \emph{a single} \WikiEnt{} as a first argument, and a PL question that produces that \WikiEnt{} as the output answer as its second argument. E.g., \textsc{Compose}(\nl{Where was \underline{Barack Obama} born?},\nl{Who was the 44th president of the USA?}). 
The function replaces the \WikiEnt{} in the first-argument PL question with the second-argument PL question and returns the resulting PL question (\nl{Where was the 44th president of the USA born?}). 
%This is similar to function composition in CCG \cite{steedman00ccg}, or a join operation in $\lambda$-DCS \cite{liang2013lambdadcs}, where the string is a function applied to  previously-computed values. \jb{I think the last sentence can be deleted, here we focus on understanding stuff and I don't think this adds to that. I think it's possible to add a discussion on the expressivity of the language and how it related to prior work, but either in related work or after defining all the operations}
\item \textbf{\textsc{Intersect}$(\cdot, \cdot)$}: 
Takes two PL questions that return lists of more than one \WikiEnt{}, and returns their intersection as the answer. The resulting PL question is of the form \nl{PL$_1$ \underline{and} PL$_2$} omitting PL$_2$'s first word (\nl{Who was born in Hawaii \underline{and} is the parent of Sasha Obama?}).
\item \textbf{\textsc{Compare}$(\cdot, \cdot)$}: 
Takes two PL questions each returning one \WikiEnt{} that can be linked to one cell in the table, denoted by Ans$_1$, Ans$_2$.
We first choose a numeric or date column in the table, if such exists. We then compare the values of this column corresponding to the rows of Ans$_1$ and  Ans$_2$. Depending on the comparison outcome, output one of (Ans$_1$, Ans$_2$) as the operation answer. 
The PL question created is of the form \nl{What has \underline{compare-op}  \underline{numeric-column-name}, PL$_1$ \underline{or} 
PL$_2$?} omitting PL$_1$ and PL$_2$'s first word. E.g. \nl{What has \underline{most recent} \underline{creation year}, the rocket of Appolo program, \underline{or} the rocket of Gemini program?}
%we use in this work the publicly available web-based QA model by \newcite{talmor2017evaluating}, which sends questions to a search engine to retrieve snippets and extracts a set of answers from the snippets.
\end{enumerate}

%\jb{missing: What \% of the questions are cross-modal and what \% single-modality. Maybe it's later but prob,. good to already have here.} \at{maybe we will summarise this at the end of 2 for lack of space?}

\textbf{2.5 Paraphrasing using AMT}
We used English-speaking AMT workers to paraphrase automatically-generated PL questions into natural language (NL). Each question was paraphrased by 1 worker and validated by 1-3 other workers. 
To avoid annotator bias \citep{geva2019we}, the number of annotators who worked on both the training and evaluation set was kept to a minimum.
We also deployed a feedback mechanism, where workers receive a bonus if a baseline model correctly answered the question after their first paraphrasing attempt, but incorrectly after they refined the paraphrase.
See supp. material for print-screens of the AMT annotator interface.
%\footnote{see supplementary material for print-screens of the AMT annotator UI}

To generate diversity, workers got a bonus if the normalized edit distance of a paraphrase compared to the PL question was higher than 0.7. A total of 971 workers were involved, and 29,918 examples were produced with an average cost of 0.33\$ per question. We split the dataset into 23,817 training, 2,441 development (dev.), and 3,660 test set examples. Context components in the dev. and test sets are disjoint, and were constructed from a disjoint set of single-modality questions.   

A shortcoming of our method for automatically generating examples is that the question distribution does not come from  a ``natural'' source. 
We argue that developing models that are capable of performing reasoning over multiple modalities is an important direction and \MMQA{} provides an opportunity to develop and evaluate such models. Moreover, this method allows to control the compositional questions created,  proving effective in creating a cheap and scalable dataset.

\textbf{2.6 Adding distractors to the context}
\bfemph{Images.}
Questions from the \textsc{\ImageListQ{}} operator 
%inherently 
require reasoning over a list of images from the same column, and hence do not require additional distractors. For \textsc{ImageQ} questions (single-image), we randomly add images that are associated with the \WikiEnts{} that appear in the table, setting a maximum of 15 distractors per question.

\bfemph{Text.}
We used DPR~\citep{karpukhin2020dense}, a neural information retrieval model, to retrieve  distractors for all questions. Each context includes exactly 10 paragraphs, where 1-2 are  gold paragraphs and the rest are distractors.
%adds to a total of $10$ paragraphs. 
Specifically, we encode the first 2 paragraphs of each Wikipedia article with the DPR encoder, and use as distractors the paragraphs with the highest dot product between their encoding and the question encoding.  
%We use the first two Wikipedia paragraphs as $>80\%$ of the answers were found in them. 
We do not allow: (a) an overlap between the distractors in the training and evaluation sets, (b) distractors originating from the gold article, (c) distractors containing an exact match to the gold answer. 
%\dl{consider moving the last 2 sentences to supp} %\jb{I did not understand which overlaps are not allowed given the above phrasing, please re-write...} \dl{organized a bit, is it clearer?}

%For the above purpose, we did not allow an overlap between the distractors in the train and evaluation sets, and also did not use distractors originating from the gold article, or containing an exact match to the gold answer. Further, we chose the first $2$ paragraphs as we observed that more than $80\%$ of the text answers, in the RC datasets, are within these paragraphs. \dl{consider removing the above if we lack space} \at{we can also say all of this much shorter}

To summarize, each of our examples contains a question, an answer, the formal representation of the PL question (ignored by our models), 
%\gi{again, is this defined somewhere?}, 
and all distractors and gold context for all modalities. This renders \MMQA{} useful for both open-domain multimodal QA, as well as context-dependant QA.

\section{Dataset Analysis}

%\at{\begin{enumerate}[topsep=0pt,itemsep=0ex,parsep=0ex]
%    \item Per compositionality type breakdown (table) + examples for each (most important)
%    \item Context type breakdown (filmography / sports / other categories)
%    \item General stats table (average number of answers / questions lengths etc ... )
%    \item -------- Submission to Supp line ------------
%    \item Majority class for multi-choice / binary questions? 
%\end{enumerate}}

%\at{TODO some intro sentence for this section}

%JB: for brevity
%\MMQA{} capitalizes on contexts from Wikipedia, that are used to automatically generate pseudo-language complex multimodal questions, which we ask crowd workers to paraphrase into natural language. 
To highlight the diversity of \MMQA{} we analyze its key statistics, domains, and lexical richness.  
%where each context contains an anchor table to which we connect images and text questions via the Wikipedia links of the table. 
%We use the contexts to automatically generate pseudo language multimodal questions, and then ask AMT workers to paraphrase the questions into natural spoken language.

\begin{wraptable}[14]{R}{0.4\columnwidth}
    \vspace{-16pt}
    \begin{center}
    \resizebox{0.4\columnwidth}{!}{
        \begin{tabular}{ll} \toprule
        {\emph{Measurement}} & {\emph{Value}} \\ \midrule
        \# Distinct Questions & 29,918 \\
        Train multimodal questions & 34.6\% \\
        Dev.+test multimodal questions & 40.1\% \\
        Train compositional questions & 58.8\% \\
        Dev.+test compositional questions & 62.3\% \\
        Average question length (words) & 18.2 \\
        %Long questions (more than 20 words) & 29\% \\
        Average \# of answers per question & 1.16 \\
        List answers & 7.4\%  \\
        %Average \# of answers for partial questions & 1.53 \\
        List answers per intermediate question & 18.9\%  \\
        Average answer length (words) & 2.1 \\
        %Answers with more than 1 word & 52.2\% \\
        \# of distinct words in questions & 49,649 \\
        \# of distinct words in answers & 20,820 \\ 
        \# of distinct context tables & 11,022 \\ \bottomrule
        \end{tabular}
        }
    \end{center}
    \caption{Key statistics for \MMQA{}.}
    \label{tab:key-stats}
    \vspace{16pt}
\end{wraptable}

\textbf{Key Statistics} \MMQA{} contains $29,918$ questions, and their main statistics are in Table \ref{tab:key-stats}. Since we focus on multimodality, we upsample the number of multimodal questions in the dev. and test sets compared to the training set. Also, about 60\% of the questions in \MMQA{} are compositional.
%\jb{define compositional}.
Questions are relatively long ($18.2$ words), but answers tend to be short ($2.1$ words).
%Answers tend to be short ($2.1$ words on average), but most ($52.2\%$) answers contain more than $1$ word.
The answer for each question can be a single answer or a list of answers. While list answers comprise only $7.4\%$ of the data, when considering compositional questions that contain an intermediate question within them, the proportion of list answers in intermediate questions is higher ($18.9\%$).
%The types of the answers are either strings or numbers.
%It is worth mentioning that complex questions in the dataset are compounded from several intermediate questions. 
%For example, the answer for a conjunction question is the intersection between two intermediate questions.%, such that the number of answers is smaller than the number of answers for each of the intermediate questions. 
%Hence, although the percentage of questions with list answers is small ($6.7\%$), the percentage of intermediate questions with list answers is significantly higher ($17.3\%$). 

\begin{figure}[t]
  \includegraphics[width=1.0\columnwidth]{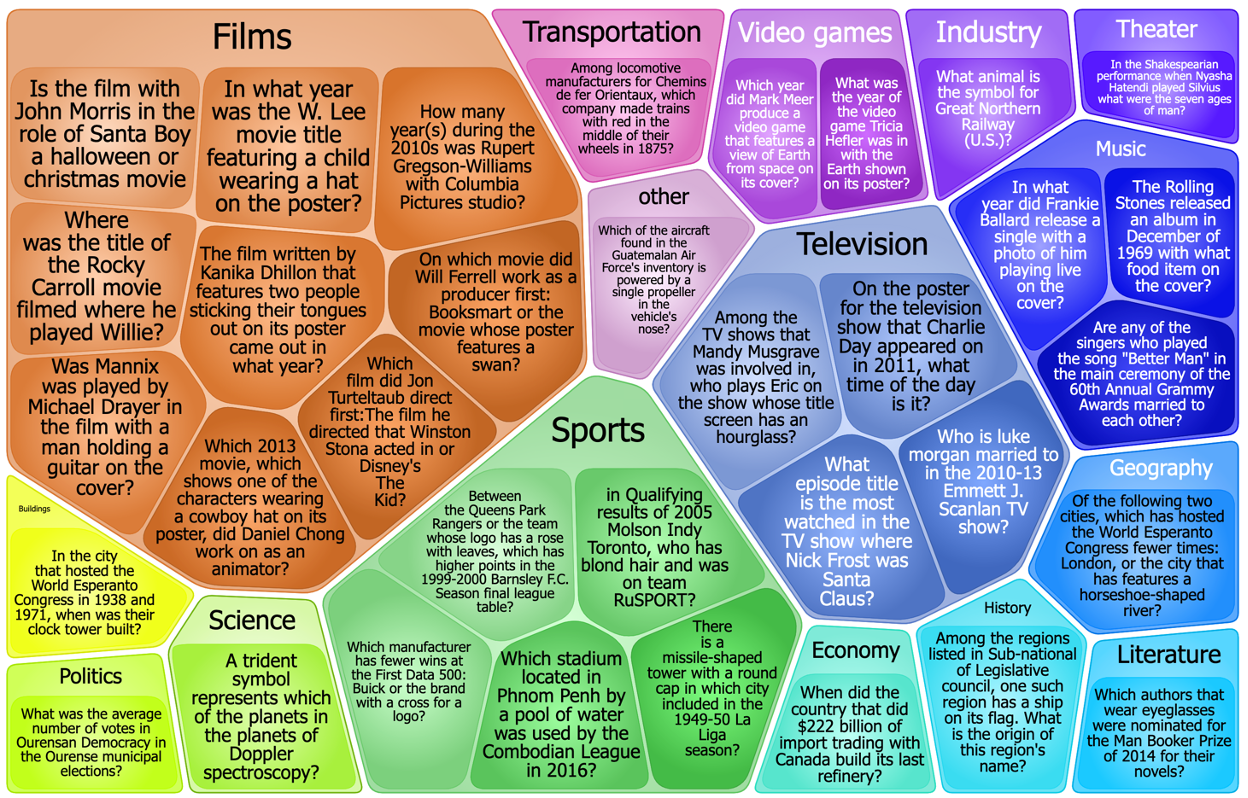}
  \caption{Domain diversity in \MMQA{}. The area of each color corresponds to the topic frequency in the dataset.}
  ~\label{fig:domain-variation-fig}
  \vspace{-10pt}
\end{figure}

%DL: we can consider to join this with the previous paragraph to save space
\textbf{Domain Diversity} Figure~\ref{fig:domain-variation-fig} shows a sample of questions from \MMQA{} categorized to different domains. While entertainment categories occupy a large portion of our dataset (Films 36\%, TV 19\%), we observe questions represent a wide variety of topics.

%\begin{wrapfigure}[14]{r}{1\columnwidth}
%  \includegraphics[width=1\columnwidth]{figures/domain_variation2.png}
%  \captionof{figure}{Domain variation}
%  ~\label{fig:domain-variation-fig}
%\end{wrapfigure}

%\begin{wrapfigure}[9]{r}{0.3\columnwidth}
%    \vspace{-14pt}
%  \includegraphics[width=0.3\columnwidth]{figures/edit_distance_hist.png}
%  \captionof{figure}{Distribution of Normalized edit distance between MG and NL questions.}
%  ~\label{fig:edit-distance-fig}
%  \vspace{14pt}
%\end{wrapfigure}

\textbf{Lexical Richness} Workers received a bonus when substantially modifying the PL questions. We observe that the average normalized edit distance between the NL questions and the PL questions is high ($0.7$),
that NL questions
%the NL questions have a high average normalized edit distance (ED) vis-a-vis the MG ones 
%, figure \ref{fig:edit-distance-fig}),
%(Avg. ED: 0.7, Std. ED: 0.15), 
are shorter (avg. length of $20.02$ vs. $22.16$ words for PL questions), 
%(Avg. word length NL: 18.63, Avg. word length MG: 22.48), 
%have a higher length variety (std. of $6.17$ vs. $4.87$ words)
%(Std. word length NL: 6.17, Std. word length MG: 4.87),
and use a richer vocabulary (\#unique words $39,319$ vs $37,108$).
%(NL \# unique words: 36,603, MG \# unique words: 33,835).

\section{Models}
Here we present our baseline models. 
%to answer multimodal questions. 
We first train models that interact with a single modality given a question (\S4.1), and
%then 
use those as building blocks in our multimodal approaches  (\S4.2). We denote the question by $Q$, context paragraphs by $\mathcal{P}$, Table by $T$ and 
context images by $\mathcal{I}$.

%In this section, we first describe our base modules capable of handling textual, tabular and visual contexts in Section 4.1, and then introduce pipeline methods to integrate them in Section 4.2. Training details for all the models are described in Section 4.3.
\vspace{-3pt}
\subsection{Single-Modality QA Modules} 
%\hfill \break 
\label{sec:single_mod}

\textbf{Text QA Module}
%In our task, multiple documents including distractors and supporting context are given, and a system needs to find and reason about the correct supporting context.
%\as{red}{
Following prior work~\citep{Min2019CompositionalQD,Asai2020LearningTR}, our text QA module takes as input a question $Q$ and a paragraph $p$ $\in \mathcal{P}$ and answers $Q$ by selecting a span in each paragraph $p$ independently, predicting the start and end positions~\citep{devlin2019bert}. Additionally,  the model returns four scores for for every paragraph $p$ corresponding to: if the answer is (\textit{i}) a span in $p$; (\textit{ii}) \nl{yes}; (\textit{iii}) \nl{no}; or (\textit{iv} ) not in $p$. 
At inference time, the model selects the paragraph that has the lowest  score for \textit{iv} -- the answer is not the paragraph. Our model is based on a pre-trained RoBERTa-large model~\citep{liu2019roberta}, fine-tuned on \MMQA{}.

% In our task, multiple documents including distractors and supporting context are given, and a system needs to find and reason about the correct supporting context. Following prior work on multi-hop question answering~\cite{Min2019CompositionalQD,Asai2020LearningTR}, our text QA model is based on pre-trained contextualized representations (i.e., RoBERTa;\citealt{liu2019roberta} ), and scores and answers each paragraph independently. 
% The model takes a question $Q = [1_1 , \ldots, q_l]$ and a paragraph $P_j = [p_1, \ldots , p_m ]$, where $m$ is the length of the question and $n$ is a length of the $j$-th paragraph in the $n$ document document collections associated with the question \at{n is used twice here, document is repeated, typo? }, $\mathbf{P} = [P_1, \ldots, P_j \ldots, P_n]$. 
% We encode the question and single paragraph jointly by concatenating the inputs with a [CLS] token and separation tokens [SEP], $u_{S} = \mathrm{RoBERTa}_\mathrm{[CLS]}(Q, P) \in\mathbb{R}^{D}$, where $D$ is the hidden dimension of pre-trained representations. Then, we calculate the probabilities of the answer being a span in $P_j$, yes, no, or not included in $P_j$, $y_{span}, y_{yes}, y_{no}, y_{no\_ans}$ as follows:
% \begin{equation}
%   [y_{span}; y_{yes}; y_{no}; y_{no\_ans}] =  \sigma(w_{label} \cdot u_{s}),
% \end{equation}
% where $w_{label} \in \mathbb{R}^{D \times 4}$ is a weight vector.
% The final answer is obtained by selecting an span or yes/no answers from the paragraph with the lowest $y_{no\_ans}$ score. 

\textbf{Table QA Module}
Following prior work \citep{herzig2020tapas}, our table QA module takes as input the question $Q$ and the table $T$, and selects a subset of the table cells and an aggregation operation to compute the final answer.
%Following prior work \citep{herzig2020tapas}, our model selects a subset of  table cells and predict  aggregation operations to compute the final answer. 
Specifically, we linearize the table $T$ by rows, with column names prepended to the corresponding cells \citep{chen2019tabfact}.  
For example, this converts the table in Figure \ref{fig:intro-fig} to the following text: \nl{Row 1: year is 1957; title is a dangerous age; role is David. Row 2...}. Next, we concatenate the question to the linearized table, and encode them using RoBERTa-large. We then pass the contextualized representation of every  token in the table cell to a  linear classifier that computes the probability of the token being selected. The score for a cell is the average of its tokens. Cells with probability $>0.5$ are selected. Finally, another linear multi-class classifier %on top of the sentence-level encoding ([CLS])
predicts an aggregation operation from \texttt{SUM}, \texttt{MEAN}, \texttt{COUNT}, \texttt{YES}, \texttt{NO}, and \texttt{NONE}. Aggregation operations are applied on the selected cells,  \texttt{YES} and \texttt{NO} operations output \nl{yes} or \nl{no}, and the \texttt{NONE} operation outputs all selected cells.

\textbf{Image QA Module}
Questions with visual information are handled by a multimodal transformer that processes the text question and pre-computed image features. 
For a question $Q$ and a set of images $\mathcal{I}$, we feed the model the question and the visual features $\Phi(i)$ extracted from each image $i \in \mathcal{I}$, along with the name of the \WikiEnt{} associated with the image.
For each image and question, the model predicts an answer from a fixed vocabulary determined by the answers in the training set and 3 special tokens: $y_{\text{dtr}}$, $y_\text{p}$ and $y_\text{n}$. In questions where the expected answer is a phrase (e.g., ImageQ), we return the answer from the image where  $p(y_{\text{dtr}})$ is lowest (similar to text QA). In questions where the expected answer is a subset of the images (e.g., Compose(TableQ,ImageQ)), we return all images where $p(y_\text{p}) > p(y_\text{n})$. Our model is based on the pre-trained model VILBERT-MT\footnote{The multi-task version of VILBERT is used, since it was shown in \cite{lu202012} that fine-tuning task-specific models from the multi-task model is generally beneficial for performance on single tasks.} \citep{lu202012}. Visual features are extracted by a vision network $\Phi$, comprised of a Faster R-CNN \citep{ren2015faster}
%with a ResNet-101 backbone \citep{he2016deep}
 pre-trained on Visual Genome \citep{krishna2017visual}. %, as in \citet{anderson2018bottom}.
%Given the high memory and computational requirements of processing multiple images, we process each image separately.

%\textcolor{red}{
%The answer $y_{dtr}$ is used for distractor images that are unrelated to the question, while $y_p$ and $y_n$ are used for images that should be or not included in the answer the model gives. At inference time, the answers from all images associated with a question are combined based on the model's predictions. For questions where a word or expression is expected (e.g. QImage), we select the answer associated with the image the model assigns the smallest probability to $y_{dtr}$. For questions where the expected answer is a subset of the input images (e.g. Compose(QImage,QTable)), we select those where the model assigns a higher probability to $y_{p}$ than to $y_{n}$, limiting the number of selected images if appropriate. 
%} \at{can we please shorten the text mark by Akari in red?}

\vspace{-3pt}
\subsection{Multimodality QA models} 
%\hfill \break 
\label{sec:pipeline}

% Having described our base modules capable of handling textual, tabular and visual contexts, we now introduce two pipeline methods to integrate them.

% In the following, we describe our {\bf single-hop auto routing} and {\bf multi-hop implicit decomposition} models. Although Both of the models are designed to answer a question given information from three modalities. We first describe our {\bf question type classification} model, only {\bf multi-hop implicit decomposition} could aggregate the information across different modalities. We first describe our {\bf Question Type Classification} model, which is used by two baselines to identify the reasoning skill the model should perform. \yw{Describe the question type classification model in two sentences.}

% Yizhong: this paragraph seems not necessary, and we can shorten it to 1-2 sentences and put this into appendix.
% \paragraph{Question type classification}
% Our question type classification model is a RoBERTa-based $k$-way (e.g., $k=24$ in our dataset) classification model to predict the target question type given a question only. 
% The model takes a question $Q$, and predicts the probability of a question being a question type $t_l$ as: $[t_1, \ldots , t_l, \ldots , t_{24}] =  \sigma(w_{type} \cdot u_{q})$,
% where $w_{type} \in \mathbb{R}^{D \times 24}$ is a weight vector, and the question type is determined with the one with the highest probability. 

\begin{wrapfigure}[18]{R}{0.37\columnwidth} 
\vspace{-7pt}
  \includegraphics[width=0.37\columnwidth]{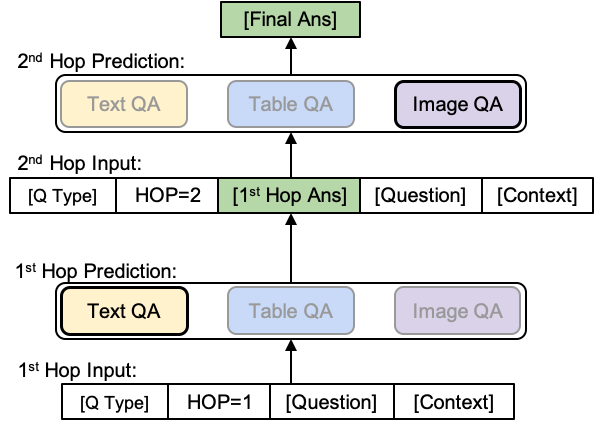}
  \captionof{figure}{\ImplicitDecomp{}: Modules with the same color share parameters. In this example, the text QA module is activated to produce the 1st-hop answer, and this intermediate answer is fed into the Image QA model to produce the final answer. Question type ([Q Type]) is determined by a separate classifier. 
  %\hh{the figure is mainly in front of the single-hop paragraph.... make sure it sits in front of the multimodal}
  }
  ~\label{fig:implicit-decomp-fig}
\vspace{7pt}
\end{wrapfigure}

We turn to models that interact with multiple modalities.

\textbf{Multi-Hop Implicit Decomposition (\ImplicitDecomp{})}
Our dataset is designed to test reasoning across modalities. As a first attempt towards this goal, we introduce a 2-hop implicit decomposition baseline, capable of combining information scattered across modalities (illustrated in Figure \ref{fig:implicit-decomp-fig}).

We first train a question-type classifier, based on RoBERTa-large, that takes a question $Q$ as input, and predicts one of the 16 possible question types (Table \ref{tab:compositionality-breakdown}). The question type can be viewed as a program, specifying the relevant modalities, their order, and the logical operations.
For example, if the question type is \textit{Compose(TextQ,TableQ)}, the first hop should be conducted on the table $T$, and the second hop on the paragraphs $\mathcal{P}$. 
In each hop, we feed the model with the question $Q$, the question type, the hop number, and the context of the corresponding modality. The model  automatically identifies which part of the question is relevant at the current hop and does not explicitly decompose the question into sub-questions  (hence the name \emph{implicit decomposition}).
In the second hop, answers from the first hop are also given as input so that the model can leverage this information and conduct cross-modal reasoning to output the final answer.  For all single-modality question types (such as \textit{TextQ} and \textit{TableQ}), the model uses only the first hop to get the answer.

\textbf{Single-Hop Routing (\AutoRouting{})} 
%\hh{propose bringing implicit decomposition before single-hop routing; right now implicit decomp is lightly explained whereas that's the interesting baseline. You can describe classification, etc in the impl. decomp and then single-hop routing explanation can be light.}
A simple approach for answering questions without cross-modal reasoning is to first determine the modality where the answer is expected to occur, and then run the corresponding single-modality module.
%Given the models in \S\ref{sec:single_mod}, one simple approach to answer a question possibly regarding arbitrary modalities is to first determine which model is best equipped to answer it, then route the question to the model. In this spirit, 
We use the aforementioned question type classifier to determine the modality where the answer will appear, route the question and the context for the predicted modality into the corresponding module, 
%Similar to \ImplicitDecomp{}, \AutoRouting{} uses a classifier that predicts the question type. The question type specifies the number of hops and the relevant modality in each hop. We feed the question and the context of the target modality into the corresponding model, 
and use the output as the final answer.
%\jb{make sure to present the result of the quest type classifier in the next section}
%\hh{I think this method extends the approach of ManyModalQa, right? if yes, worth mentioning.}

% Single-hop Auto Routing baseline we first run the question type classification model described below to find from which modality the answer should be found (e.g., if the question is compositional from table to text, the final answer should be extracted from $\mathbb{P}$). 
% Once it determines the target modality, it runs one of the Only-text, Only-table or Only-Image to find the final answer from the target modality. 

\textbf{Question-only and Context-only baselines} 
\if0{
To ensure that our dataset is free of simple shortcuts that can answer the question without knowing either the context or the question, we set up a question-only baseline and a context-only baseline for sanity check, as is suggested by \cite{kaushik2018much}. 
Different from the original paper, for question-only baseline, we employ a seq-to-seq architecture to generate the answer given only the question. We finetuned the BART-large~\citep{bart} model on our question-answer pairs. This is also similar to the closed-book setting of open-domain quetion answering~\citep{raffel2019exploring}. 
For the context-only baseline, we use the same pipeline as \AutoRouting{}, but void the question when each single-modality model is used to produce the answer. 
}\fi
We run the question-only and context-only baselines, suggested by \citet{kaushik2018much}. Our question-only baseline is BART-large~\citep{bart}: a sequence-to-sequence model that directly generates the answer given the question. 
%The model is fined-tuned on our question-answer pairs, similar to similar to the closed-book setting of open-domain question answering~\citep{raffel2019exploring}. 
For the context-only baseline, we first predict the question type using the classifier described above to pick a target module. We then feed the relevant context to the target module, replacing the question with an empty string. 
% \jb{how do we choose a modality here?}

% \paragraph{Question Only}
% In order to evaluate whether the additional table, text, and image contexts are required to correctly answer the questions, we train a BART-large~\citep{bart} generative model on question-answer pairs. Given a question-answer pair, the input to the model is the encoded question and the model's loss function is the cross-entropy between the model's output and the question's answer. Because BART is a Seq-to-Seq model, we recompose list answers to sequences such that each sequence includes the list answers separated by a separating character. \oy{are the last two parts over-descriptive? can we stay with the first line?}

% Our question only baseline is a BART-large~\citep{bart} model, which receives the questions without additional contexts. The low performance of the model (Dev F1: 0.16, Dev EM:0.138) suggests that the majority of the questions require reasoning over the context data. When analysing the successful predictions of the model, we notice that 48\% of the questions the model answers correctly require either boolean or comparison reasoning. In addition, the model achieves better results on uni-modal questions (EM:16.1) than multimodal ones (EM:11.3). Hence, we suggest that some of the success of the model resides in the possibility of estimating the answer based on the distribution of question-answer pairs seen during training. \oy{not sure about the last sentence, we can also elaborate about answers that are colors or numbers, but then we'll have to explain more notations}

\vspace{-3pt}
\subsection{Training and Supervision}

%\hfill \break 
\label{sec:training}
Our dataset provides rich supervision including not only the final answer but also question types and intermediate results. Therefore, we can train the pipeline modules in a supervised fashion. Specifically, we train the question type classifier using a cross entropy loss w.r.t the gold question type. For \AutoRouting{}, each QA module is trained with the subset of samples whose final answer can be extracted from the corresponding modality. For \ImplicitDecomp{}, only one model is trained per modality, which is used to answer both the first-hop and second-hop questions. The question-only and context-only baselines are trained in the corresponding format.

\section{Experiments}

We evaluate models in three different setups: 
%(1) the subset of questions for which the answer is in the text, table or image modality respectively (e.g. \emph{Ans. in Image}); 
(1) questions that require a single modality to answer (\emph{Single Modality}); (2) questions that require reasoning over multiple modalities (\emph{Multi Modality}); (3) and all questions (\emph{All}). 
Our evaluation metrics need to support lists of answers, and thus we use average F$_1$ and Exact Match (EM), as described in \cite{dua2019drop}, where answers on the gold and predicted lists are aligned. 
Human performance is estimated with $9$ expert annotators, who answered 145 questions. Test results are 
are reported using a single run (one random seed).  
%We find that a 2-hop version of the \ImplicitDecomp{} model was sufficient to answer all question types.  
\begin{wraptable}[9]{R}{0.57\columnwidth}
\vspace{-14pt}
\begin{center}
\resizebox{0.57\columnwidth}{!}{
\begin{tabular}{l|cc|cc|cc}
                            & \multicolumn{2}{c|}{Single Modality} & \multicolumn{2}{c|}{Mutli Modality} & \multicolumn{2}{c}{All} \\ \hline
                            & EM          & F$_1$  & EM          & F$_1$  & EM          & F$_1$  \\ \hline
Question-only\footnotemark[2]               &  14.2 & 17.0 & 16.9 & 19.5 & 15.3 & 18.0 \\ 
Context-only                & 8.0 & 10.2 & 6.6 & 8.5 & 7.4 & 9.5 \\ 
\hline
\AutoRouting{}              &  48.9 & 57.1 & 32.0 & 38.2 & 42.1 & 49.5 \\ 
\ImplicitDecomp{}           & \textbf{51.1} & \textbf{58.8} & \textbf{46.5} & \textbf{51.7} & \textbf{49.3} & \textbf{55.9}  \\ 
\hline \hline
%\hhline{=|==|==|==}
Human             & 87.9 & 92.5 & 84.8 & 90.1 & 86.2 & 91.2 \\ 
%\hline
\end{tabular}}
\end{center}
\caption{Test set results }
\label{tab:results}
\vspace{14pt}
\end{wraptable}

We show results in Table~\ref{tab:results}.\footnote{
We conjecture that the higher performance exhibited by the question-only baseline compared to the context-only baseline is due to the fact that in comparison questions the model needs to choose one answer from two candidates, of which one usually appears in the question, allowing the model to obtain 50\% accuracy by guessing.}
\ImplicitDecomp{} achieves significantly higher performance (55.9 F$_1$) compared to the other baselines, but lower than human performance (91.2 F$_1$ with provided context, and 84.8 F$_1$ in the open-domain setting over all of Wikipedia), suggesting ample room for improvement. 
On the \emph{Multi Modality} subset, \ImplicitDecomp{} substantially improves performance compared to \AutoRouting{} ($38.2 \rightarrow 51.7$), emphasizing the superiority of our approach on multi-hop questions, while on single-hop questions this gap is smaller.

Since automatic evaluation of performance is non-trivial in our setup, we also manually evaluate human performance. In 94.5\% of the cases, answers are either identical or semantically equivalent to the gold answer, 0.7\% have an error in the question, and 4.8\% are human errors. Human errors are owing to the length of the context, resulting in human fatigue (which models do not suffer from).

\textbf{Analysis} To demonstrate that \ImplicitDecomp{} indeed performs multi-hop reasoning, successfully answering intermediate questions, we analyze \ImplicitDecomp{} predictions for multimodal questions generated using the \emph{Compose}, \emph{Compare} and \emph{Intersect} operations (Table~\ref{tab:predictions_analysis}). 
For these questions, we find that when the 1st-hop answer is correct, the model achieves an F$_1$ of 63.9, whereas when the 1st-hop prediction is incorrect, the F$_1$ drops to 37.4. This suggests that the model relies on the 1st-hop answer, effectively performing multi-hop reasoning. Last, our question type classifier obtains a high accuracy of 91.5\% on the test set.

\begin{table}[t]
\centering
\resizebox{1.0\textwidth}{!}{
\begin{tabular}{l|l|l|l|c|c}
 %\toprule
 \textbf{Type} & \textbf{Question} & \textbf{1st-hop prediction} & \textbf{Final prediction} & \textbf{1st-hop F$_1$} & \textbf{F$_1$} \\ 
\hline
 Compose &\emph{What part did Kym Karath play in the TV show } &
 Lassie & Kathy Vaughn  & 62.3 & 50.8 \\ 
 & \emph{ whose poster features a dog?} &   &  &  &  \\ 

\hline
 Compare & \emph{Which video game was Wes Johnson involved in earlier:} & Hammer \& Sickle & Hammer \& Sickle & 55.7 & 61.1  \\
 & \emph{Fallout 4 or the game whose cover shows a gun-wielding man?} &  &  & &  \\
\hline
 Intersect & \emph{Which album, released in December of 2011, has a man wearing} & TY.O, & Back to Love & 33.5 &  55.1 \\
 & \emph{sunglasses on its cover, and was released under the RCA label?}  &
 Back to Love &  &  &  \\
\end{tabular}}
\caption{Examples where \ImplicitDecomp{} correctly answers both the intermediate and the entire question, and a breakdown of the 1st-hop F$_1$ and final F$_1$ for the three logical operations: Compose, Compare, and Intersect.}
\label{tab:predictions_analysis}
\vspace{-5pt}
\end{table}

To test whether the compositional questions created are indeed multi-hop, we conducted a qualitative analysis over 50 questions as suggested by \citep{min2019compositional}.
We find that 6\% are of the \emph{Weak Distractors} category, that is, questions such as \nl{What year... } when there is only one year appearing in the context, making the question easy. 2\% have \emph{Redundant evidence}, that is, questions such as \nl{Which Donald Trump TV show has ...} where there is only one TV show starring Donald Trump, making the rest of the question redundant. The remaining 92\% indeed require multi-hop reasoning.

\comment{
\at{\begin{enumerate}[topsep=0pt,itemsep=0ex,parsep=0ex]
    \item Eval description, etc ... 
    \item \textbf{Baseline results table}, results per baseline (see baseline models) + HUMAN EVAL
    \item \textbf{Per question-type accuracy breakdown table} + error analysis in text
    \item "Sewon Min - is multi-hop question / Compositional questions do not necessitate multi-hop reasoning" analysis and breakdown
    \cite{Min2019CompositionalQD}
    \item autorouting analysis, and perhaps questions of good and bad programs
    \item -------- Submission to Supp line ------------
    \item Distractor analysis? 
\end{enumerate}}
}
\section{Related Work}

Visual question answering---i.e., the task of answering questions about images---has been widely explored in previous work \citep{antol2015vqa, balanced_binary_vqa, balanced_vqa_v2,johnson2017clevr,hudson2019gqa,zellers2019recognition,singh2019towards,methani2020plotqa}, ranging from synthetic images to scientific plots. Our work differs significantly from those, by including more complex, multi-hop questions that require reasoning over text, tables and images. Currently, the most successful paradigm in VQA is fine-tuning models pre-trained on large amounts of image captioning data \citep{tan-bansal-2019-lxmert,lu2019vilbert,lu202012,su2019vl,chen2019uniter,li2020oscar}, an approach we follow for answering image-related questions.

% Thanks for this idea. We tested zero-shot question type classification performance of our BERT-based modality classification model on ManyModalQA, a collection of single-hop questions across three modalities. As our original question classification model is a 16-way classification, we mapped the 16 types into 3 types based on the modality from which the final answer would be extracted (e.g., Compose(text, table) --> text). The zero-shot accuracy on ManyModalQA is 65.39 %., whereas the original paper’s classification results after fine-tuning is 77.93%. For the final version, we will also fine-tune on ManyModalQA and compare our classifier accuracy.

\textsc{ManyModalQA} \citep{hannan2020manymodalqa} move beyond directing the question to an image-only context, to choosing between an image, a text, and a table. Their work focuses on routing the question to the correct context modality. Our question-type classifier, based on RoBERTa-large reaches an accuracy of 91.4\% on our 16 possible question types, showing that the main challenge in \textsc{MMQA} is reasoning over the context rather than identifying the question type.
%and 65.39\% in zero-shot when classifying the 3 major modalities in \textsc{ManyModalQA}, compared to the original paper’s classification results after fine-tuning is 77.93\%.
%With 30K examples, our dataset is also 3x larger than \textsc{ManyModalQA}.

\textsc{HybridQA} \citep{chen2020hybridqa} presents a cross-modality reasoning challenge over tabular and textual data. A fundamental difference is that our setup offers cross-modality reasoning over images as well. In addition, our approach is cheaper to annotate since it requires only paraphrasing, and the question type distribution is more controllable (we offer 16 major question types vs. 6 in \textsc{HybridQA}). Moreover, our text passages are chosen using the question, answer and table, while in \textsc{HybridQA} only WikiEntities from the table are used to find text passages.

The model proposed in \textsc{HybridQA} introduces a
heuristic for linking the text passage to the table cells, which may lead to performance degradation. Conversely, our model uses (automatically-annotated) intermediate multi-hop answers, to perform reasoning and linking implicitly over the full table and text, which should lead to more robust reasoning, in particular when reasoning over multiple table cells, as well as for narrative tracking and co-reference over the full text.  
%In parallel to us a new de-contextualised variant of HybridQA has been composed \citep{chen2020open} with includes an open domain setup similar to ours.
%linking step heuristic for passage and cell retrieval, that are performed explicitly alongside the reasoning step, and may pose a potential performance weakness. Our setup, although aided by intermediate multi-hop answers, performs the reasoning and linking implicitly over the full table and text, an approach that is more robust, in particular to reasoning over multiple table cells as well as narrative tracking and co-reference over the full text.  
In parallel to this work, a new open-domain variant of \textsc{HybridQA} has been released by \citet{chen2020open}.

%\at{comparison to HybridQA, manymodalQA, and \citep{chen2020open}}

%\at{TODO: Gabriel please add related work about VQA here... Reviewer4 \nl{The dataset would be a useful resource for multimodal QA advancement but the manuscript in the current form disregards all of the prior work that has happened in the space including the work on VQA [1], TextVQA [2] and PlotQA[3]. There has been quite a lot of work on multimodality QA and the paper should address those and take learnings from those papers. Specifically, the model introduced in the paper is too crude and specific to the dataset generation scheme (like depending on the question type) which limits the applicability of the dataset as well as the further approaches that will be developed on this dataset. For example, M4C [4] a model on TextVQA uses transformers by projecting different modalities into the same space and concatenating them as a single sequence. Comparison against similar approaches on MultiModalQA would be more useful compared to hardcoded approaches like ImplicitDecomp. This approach reminds me of the approaches that were developed on CLEVR using program synthesis and weren’t useful in the long run. Similar to TableQA modules, tables + text + images can be input to the M4C transformer as a single sequence and the classification accuracy can be calculated.} }

\section{Conclusion}

We present \MMQA{}, a new QA dataset that contains 29,918 examples, 35.7\% of which require cross-modality reasoning. %over text, tables and images.
We describe a novel framework for generating complex multimodal questions at scale, 
%involving automatic generation of questions in pseudo-language over the contexts, and using crowd workers to paraphrase.
%We perform a detailed analysis of the dataset, which showcases its diversity and unique multimodal properties, and we extensively evaluate the results using a variety of baselines. 
and %perform detailed analysis which 
showcase the diversity and multimodal properties of the resulting dataset.
We evaluate \MMQA{} using a variety of models, and confirm that the best model exploits the multimodality of the dataset and takes into account multi-hop reasoning via implicit decomposition. However, human performance substantially exceeds the best model, establishing the need for further research involving multiple modalities in question answering systems,     
%We find that the best model by far exploits the multi-modality of the dataset.
%We hope our work will drive new research involving multiple modalities in question answering systems.
which we hope that our work will drive.
\section{Acknowledgments}
We thank our colleagues at The Allen Institute of AI, James Ferguson and Amir Globerson.
This research was partially supported by  The Blavatnik Computer Science Research Fund and The Yandex Initiative for Machine Learning,  the Edmond J. Safra Center for Bioinformatics at Tel-Aviv University, and the European Union’s Seventh Framework Programme (FP7) under grant agreement no. 802800-DELPHI.
Special thanks to Carrot Search which allowed use to use their FoamTree visualization. 

\bibliography{iclr2021_conference}
\bibliographystyle{iclr2021_conference}

\appendix
\section{Appendix - Dataset Generation}
In this section we provide more details on how we extracted tables from Wikipedia, parsed them, enriched them with images and text questions, generated single modality questions and paraphrased our machine-generated questions.

\subsection{Wikipedia tables as anchors}~\label{sec: wikipeda tables supp}

\paragraph{Extraction} As specified in the main paper we use the 01-01-2020 English Wikipedia dump of Wikipedia that contains roughly $3M$ tables. From the said dump, we extracted all tables and selected those that meet the following criteria: (a) The tables contain $10-25$ rows, i.e. they are not too short or too long; (b) At least three images are associated with the table. This results in a total of $700K$ remaining tables.

To extract tables from Wikipeida we created a customized version of the publicly available WikiTextParser package~\footnote{https://github.com/5j9/wikitextparser}. The customization process included adding the ability to extract table titles and modifying issues regarding rows and columns spans. 

\paragraph{Classifying Table Columns} 
During question generation, some methods in our logical language require certain restrictions on table columns. For example, \textsc{Compare}$(\cdot, \cdot)$ requires a numeric or date column to exist in the table. 
Hence, when we extract tables we also classify columns in the tables based on the data they incorporate. We define 3 \emph{semantic types} for columns: numeric, date and index.  
The classification is built such that: (1) a column for which all the cells can be parsed into date objects is classified as a \emph{date column}; (2) a column for which all the cells can be parsed into numeric values is classified as a \emph{numeric column}; (3) a \emph{numeric column} with consecutive values is classified as an \emph{index column}.

\subsection{Connecting Images and Text to Tables}
%\dl{If I am not mistaken we do not refer from the main to this section}

\subsubsection{Images}

%\dl{the below paragraph appears in the main in a similar fashion we can decide to cut it later}
We elaborate on two cases: (a) in-table images and (b) images from pages of linked \WikiEnts{}. 

\paragraph{In-table images}
Some tables feature images inside the table cells. These columns usually appear in tables that include lists of objects, such that another column in the table includes a description of the cells depicted in the images. For example, a Wikipedia table can list buildings of a certain city, such that one column includes the names, or \WikiEnts{}, of the buildings, and a different column would contain the buildings' images.

However, it is not given in the table which column provides the description for the images, hence we deploy a linear classifier that aims to assess that.
The features of the classifier include the column's distance from the leftmost column, the percentage of unique cells, the percentage of cells with exactly one \WikiEnt{}, the percentage of cells with short text (at most 2 chars), and the column's header.

\subparagraph{The MediaWiki API}
We use the MediaWiki API package~\footnote{https://www.mediawiki.org/wiki/API:Main\_page} to retrieve the URL of the image. We assign a unique identifier to each image based on the table, row, and the column associated with it. 
We store the images in Amazon S3 using the image's identifier.

\paragraph{Images from \WikiEnts{}} 
Although some Wikipedia tables include images, this is not the common case. More frequently, a table would contain a column of \WikiEnts{} that could potentially be linked to images. Hence, we need to retrieve and associate between \WikiEnts{} and their images.
 
To associate the \WikiEnts{} with their representative image, we first go over the entire Wikipedia dump and create a mapping between the \WikiEnts{} and their associated image Wikipedia files.
Next, we try and retrieve the images' URLs from the Wikipedia file using the MediaWiki API, similar to the in-table images. Afterwards, each image is stored in S3 with the matching entities name, and identified using the \WikiEnt{} title. Because each image is associated with a specific \WikiEnt{}, the image is cached and can be used across several contexts. 

The process of matching \WikiEnts{} to their representative image cannot be done for all entities. For several reasons: (i)
files in Wikipedia can be removed over time, which means that we will not be able to successfully retrieve an image for every mapped entity; (ii) some \WikiEnts{} do not have an image associated with them in the Wikipedia dump; (iii) Wikipedia editors do not always link table cells to \WikiEnts{}; (iv) some images are not representative of the \WikiEnts{} they depict. 
For example, certain film entities are depicted using the same film logo. This image is not descriptive of a specific film, and cannot be used during question generation. In order to avoid associating \WikiEnts{} with non-descriptive images, we created a list of frequently used non-representative images, and did not link images that were included in this list.

Overall, we obtain 57,713 images, with 889 in-table images and 56,824 \WikiEnt{} images.

\subsection{Generating Single-Modality Questions}

\subsubsection{Generating Image Questions}

\paragraph{Single Images} 
When generating single-image questions, we show Amazon Mechanical Turk (AMT) crowdworkers an image, alongside its \WikiEnt{}, and ask them to phrase a question about the image with the entity being the focus of the question.

\begin{figure}[h]
  \frame{\includegraphics[width=1.0\columnwidth]{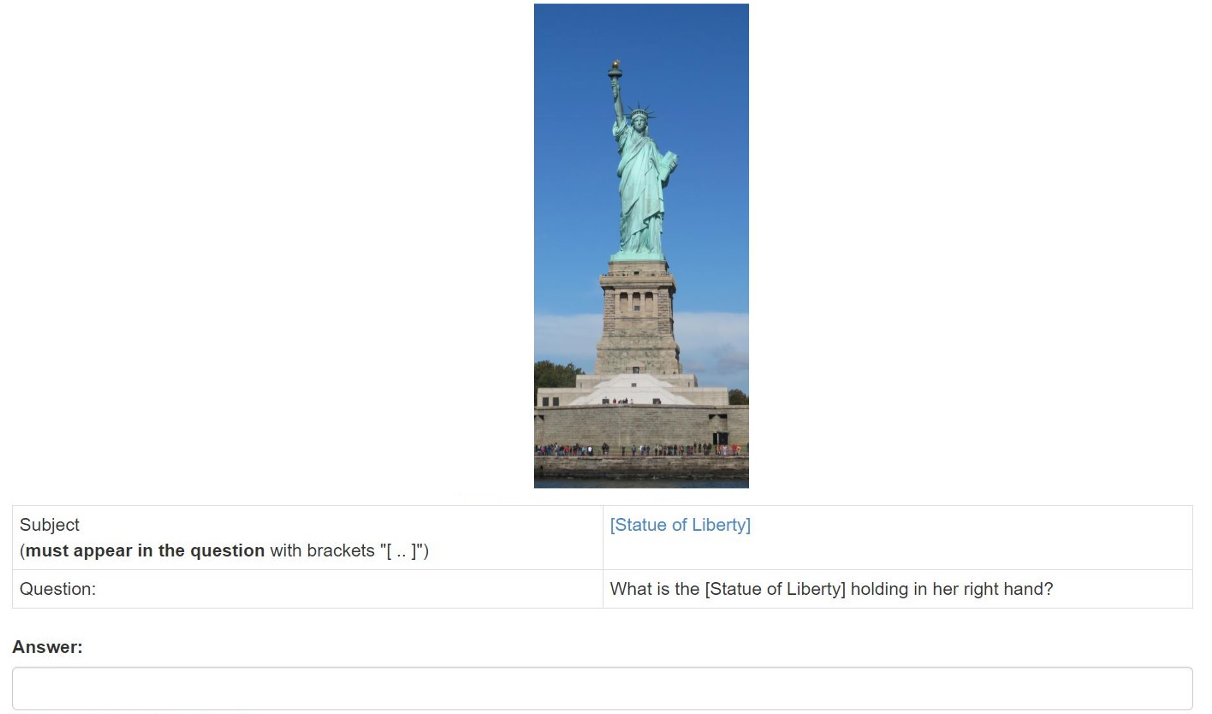}}
  \caption{When generating single-image questions, we show AMT workers an image alongside its \WikiEnt{}, and ask them to phrase a question about the image with the entity being the focus of the question.}
  ~\label{fig:amt-ui-single-image-generation}
\end{figure}

The User Interface (UI) is depicted in figure \ref{fig:amt-ui-single-image-generation}. As can be seen the AMT worker is shown an image and a \WikiEnt{} ([Statue of Liberty]), and is asked to generate an appropriate question, and its matching answer, e.g. \nl{What is the [Statue of Liberty] holding in her right hand?}. Additionally, the UI has a comments input, to communicate any thoughts, concerns or questions. We have pursued communication with AMT workers based on the comments they left us. 

To confirm that the question is suitable in an open-domain setting, we dissuade workers from asking questions about a temporary, i.e. "non-stable", features of the image. For instance, the question \nl{What is the [Statue of Liberty] holding in her right hand?} is valid in an open-domain setting, since it is unlikely that in other pictures of the the \WikiEnt{} the answer to the question would differ. However, a question such as \nl{What is the color of the sky behind the [Statue of Liberty]?} could not be used in an open-domain setting, since a picture of the same \WikiEnt{} in different times of the day would lead to a different answer.   

We note that we apply a-priori filtering on images that will be good candidates for multi-modal questions. We do so by selecting \WikiEnts{} that were associated to questions from other modalities.

Post generating the questions, we ask a separate group of AMT workers to verify that the question meets the task's criteria. 

\paragraph{Image Lists} 
For questions with a list of images, we use images that appear in the same column of a table. To generate these questions, AMT workers were given the images and asked to phrase a binary question about a distinctive feature of the entities that a subset of the images share. 

\begin{figure}[h]
  \frame{\includegraphics[width=1.0\columnwidth]{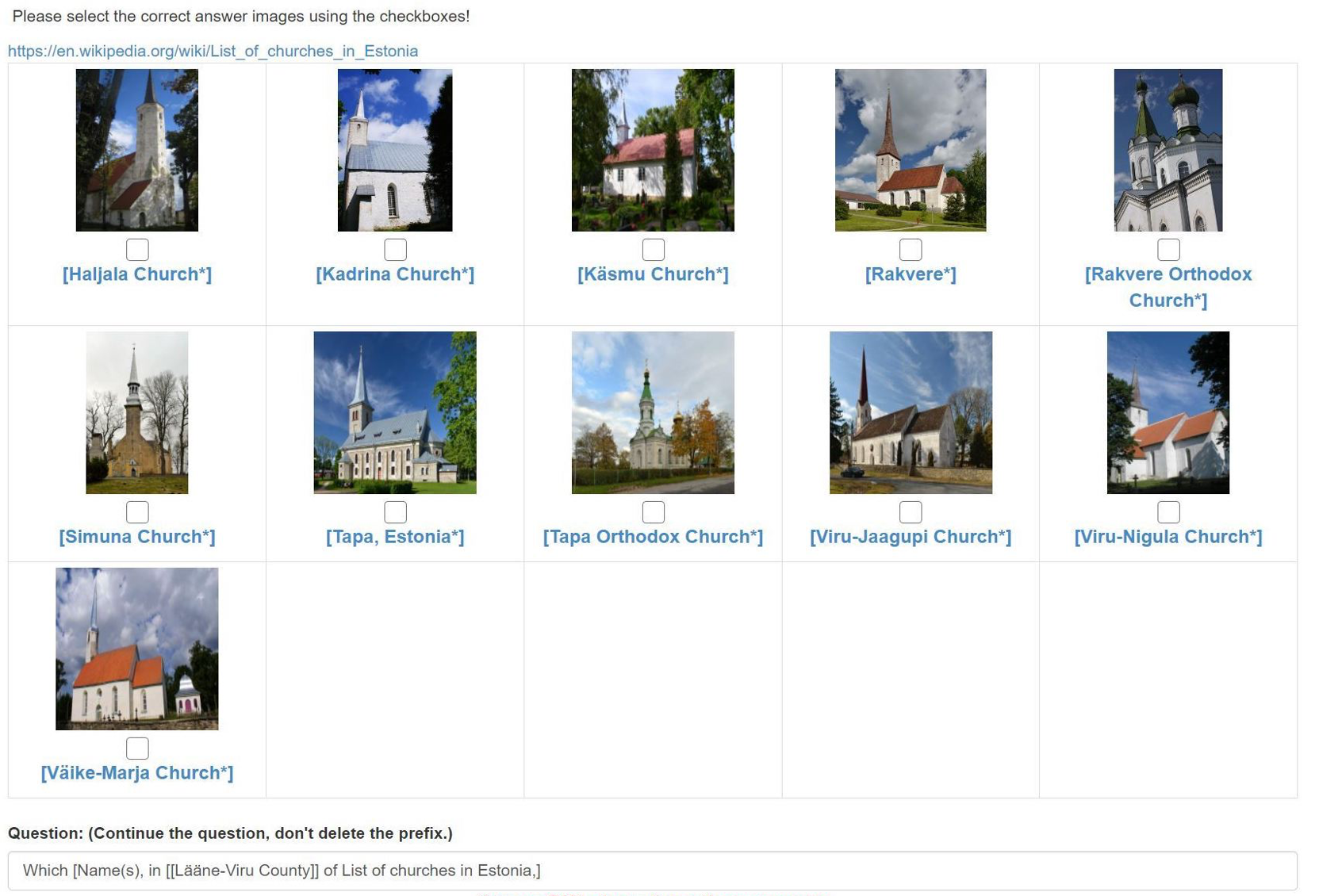}}
  \caption{The UI for our image list questions. AMT workers were given images originating from the same column, and were asked to phrase a binary question about a distinctive feature of the entities that a subset of the images share. E.g., given a list of buildings,the worker could ask \nl{Which of the buildings have a light blue roof?}}
  ~\label{fig:image-list-ui-example}
\end{figure}

The UI is depicted in figure \ref{fig:image-list-ui-example}. As can be seen the AMT worker receives a list of images from a single column, and is asked to generate a binary question that applies to all of them and provide its answer. To confirm that the question is suitable for an open-domain setting, we ask the workers to confirm that their question asks about a non-temporary, i.e. stable, feature of the images, such as geographic features or color of a person's eyes. 

For the image lists questions to be challenging, we heuristically filter columns that can be used as anchors for the image lists questions. Every such column needs to include at least 4 \WikiEnts{}, no more than 3 duplicate images, and no more than 2  \WikiEnts{} without images. 

Post generating the questions, we ask a separate group of AMT workers to verify that the question meets the generation criteria.

This process results in 2,268 single image questions and 2,223 list image questions that are later used to create multimodal questions.

\subsubsection{Generating Text Questions}
\label{sec: gen text Qs}
We leveraged various text Question Answering (QA) Reading Comprehension (RC) datasets in order to obtain single modality free text questions, that are used to compose multimodal questions (\emph{NQ} - \cite{kwiatkowski2019natural}, \emph{BoolQ} - \cite{clark2019boolq}, \emph{HotpotQA} - \cite{yang2018hotpotqa}). We describe how we linked the questions in the RC datasets to the anchor tables of our contexts.

\paragraph{Linking text questions to tables} 
In order to link between questions in the RC datasets and the tables, for each question in the RC datasets we applied the following process: (1) we extracted all the \WikiEnts{} from the HTML of the question's gold article; (2) we searched for matches between the tokens of each extracted \WikiEnt{} and the question using spaCy (\cite{spacy2}); (3) if a match was found, we marked the \WikiEnt{} as one that appears in the question; (4) if a \WikiEnt{} is shared between the question and table we connected them. 
We note that since \emph{BoolQ} and \emph{HotpotQA} do not provide the HTML of the question's context as part of the dataset, we retrieved the HTML file of each context using the MediaWiki API. 

\comment{
\dl{AT, do we want to submit this para?}
\paragraph{Processing HotpotQA} 
The plain text paragraphs provided in \emph{HotpotQA} do not match the format of \emph{BoolQ} and \emph{NQ} due to formatting reasons (e.g. special chars symbolizing italics). 
For the purpose of maintaining consistency amongst \emph{HotpotQA} and the other RC datasets, we replaced each gold paragraph in \emph{HotpotQA} with a paragraph we retrieved from the MediaWiki API. We used the edit distance measure to find the text most resembling the original. This process led us to use the 01-11-2017 Wikipedia version. To ensure that the replacement text maintains the answer-ability of the questions, we filtered out questions where: (1) the associated texts had an edit distance greater than 0.2; or (2) the text does not contain the the span of the answer. In this process, we filtered out 16,493 questions out of 90,447.
}

\setcounter{subsection}{4}
\subsection{Paraphrasing using AMT}

We asked AMT workers to paraphrase automatically-generated pseudo language (PL) questions into natural language. To do so we created a special AMT user interface, that provides various method of feedback as to the quality of paraphrasing.

\begin{figure}[t]
  \frame{\includegraphics[width=1.0\columnwidth]{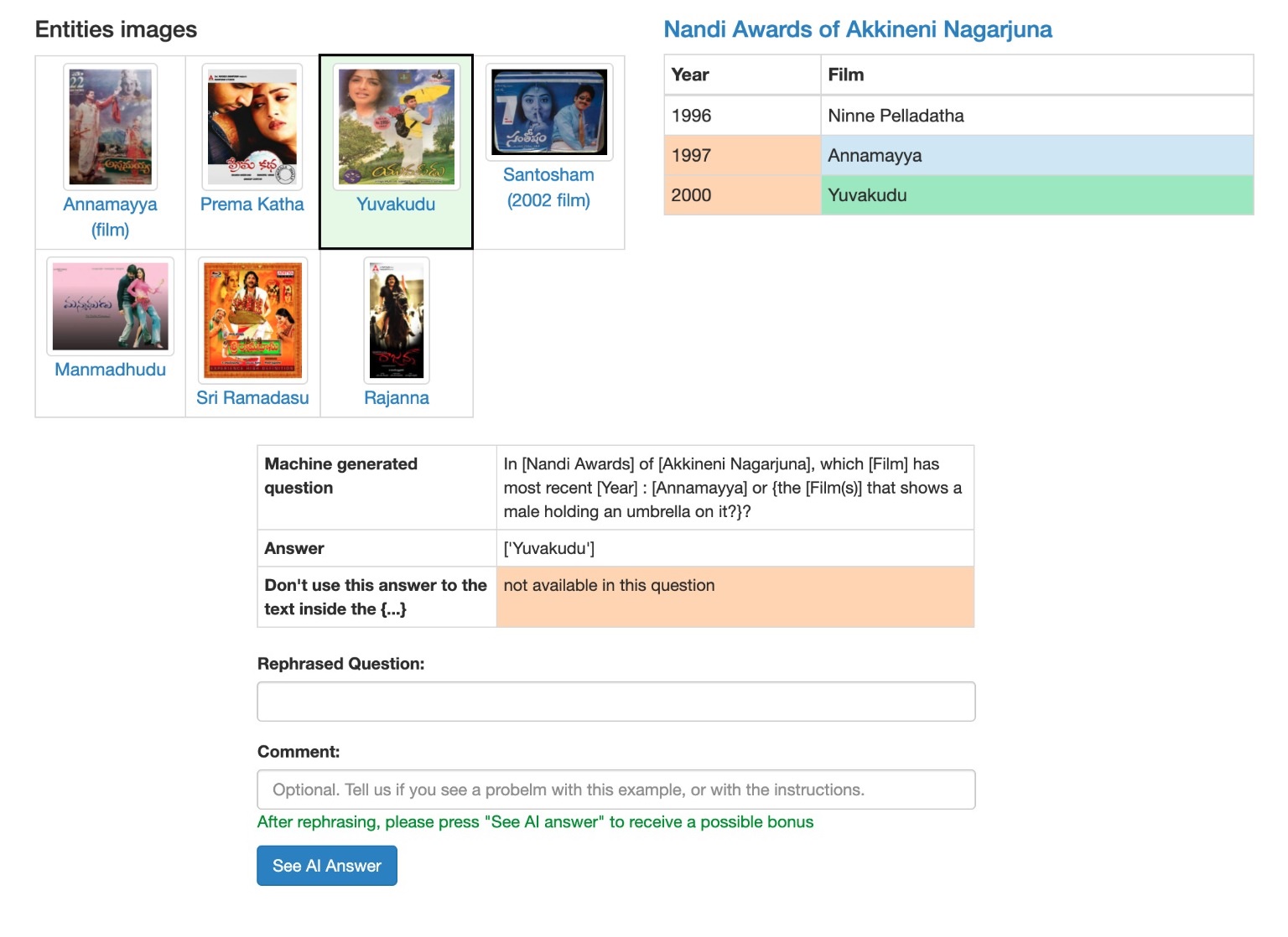}}
  \caption{An example of a PL question presented in our UI for the AMT workers during paraphrasing. For each question, we show the PL question alongside relevant context from all the modalities, and ask from the AMT workers to rephrase the PL question in NL.}
  ~\label{fig:amt-ui-rephrasing-example}
\end{figure}

The UI for the paraphrasing task is depicted in figure \ref{fig:amt-ui-rephrasing-example}.
As can be seen the AMT workers are presented with a machine generated question, its answer, the 1st hop answer (denoted \nl{Bridge answer} in the UI, presented if it exists for the question), and relevant parts of the context. The worker is asked to paraphrase the question such that it maintains its original meaning, but is phrased naturally. 

We also deployed a feedback mechanism, where workers receive a bonus if a baseline model correctly answered the question after their first paraphrasing attempt, but incorrectly after they refined the paraphrase. The workers gain access to the model feedback by pressing \nl{See AI Answer} button in the UI.
To generate diversity, workers got a bonus if the normalized edit distance of a paraphrase compared to the PL question was higher than 0.7, this feedback was presented to them automatically in each rephrase.

\begin{figure}[t]
  \frame{\includegraphics[width=1.0\columnwidth]{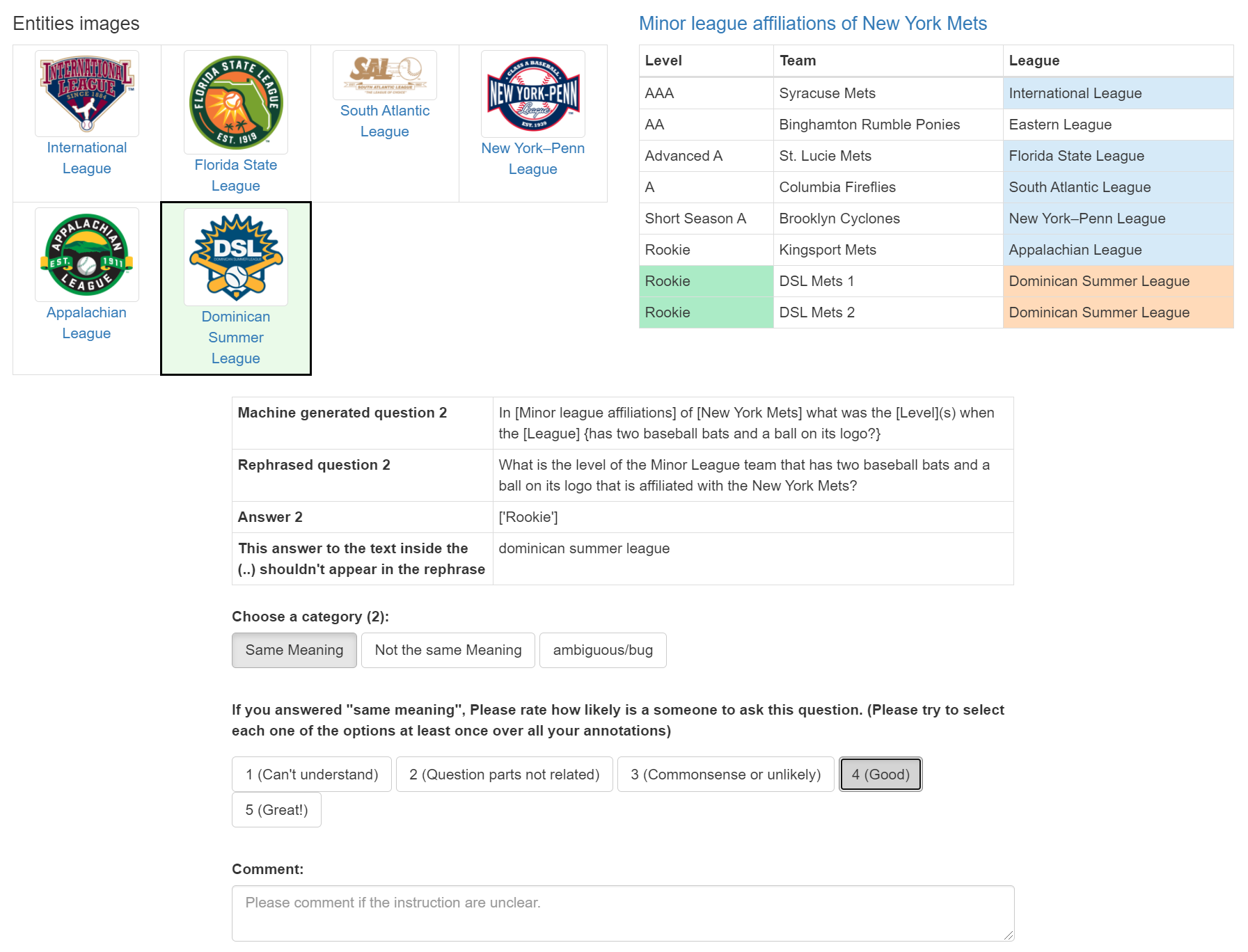}}
  \caption{An example for the rephrasing validation task in our UI for the AMT validation workers. For each rephrased question, we show an automatically-generated PL question and its rephrasing along with the relevant context from all the modalities. We ask the AMT validation workers to determine whether the rephrasing maintains the meaning of the question, and how likely it is that the rephrase question would have been asked by a human.}
  ~\label{fig:amt-ui-rephrasing-validation}
\end{figure}

Post generating the the questions, we asked a separate group of AMT workers to verify that the question meets the generation criteria. The UI for the validation task is shown in figure \ref{fig:amt-ui-rephrasing-validation}. The workers receive the original machine generated PL question, the human rephrase, and relevant parts of the context. They are tasked with determining whether the rephrase indeed keeps the same meaning, and as a another measure of quality they rate how likely it is that the rephrase question would have been asked by a human.

\section{Appendix - Dataset Analysis}

\begin{wrapfigure}[10]{R}{0.44\textwidth}
\vspace{-15pt}
  \includegraphics[width=\linewidth]{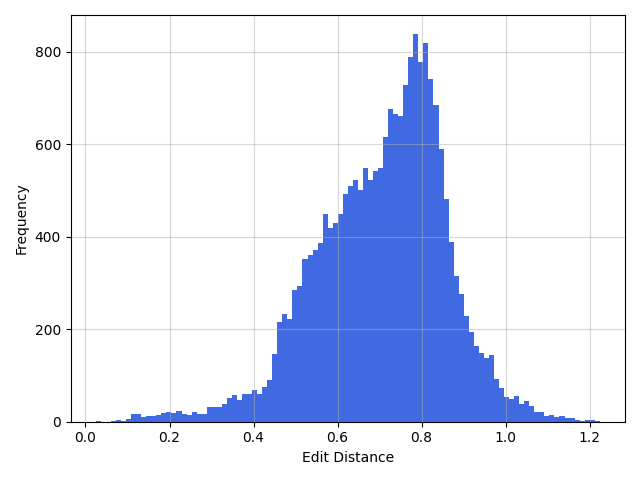}
  \captionof{figure}{MG and NL edit distance similarity.}
  ~\label{fig:edit-distance-fig}
 \vspace{15pt}
\end{wrapfigure}

\textbf{Paraphrasing Richness} Crowd workers received a bonus when substantially modifying the automatically-generated questions, to encourage diversity in the natural language questions. 
Figure \ref{fig:edit-distance-fig} shows the normalized edit distance distribution 
between the NL questions and the PL questions.

\section{Appendix - Models}
We provide additional information on some of the modules used and how we applied them.

\subsection{Single-Modality QA Modules}

\paragraph{Image QA Module} Our starting point is VILBERT-MT \citep{lu202012, lu2019vilbert}, a model trained on a wide variety of vision-and-language tasks that includes visual question answering, caption-based image retrieval, grounding referring expressions and multimodal verification. The model sees as inputs image region features extracted by a vision network (Faster R-CNN \citep{ren2015faster} with a ResNet-101 backbone \citep{he2016deep}) pre-trained on the Visual Genome dataset \citep{krishna2017visual}, as in \citet{anderson2018bottom}. From each image, we extracted a 100 2048-dimensional region features. Given the high memory and computational requirements of processing multiple images, we processed each image separately, and combined the answers as described in the main paper.

\section{Appendix - Experiments}

\begin{figure}[h]
  \includegraphics[width=1.0\columnwidth]{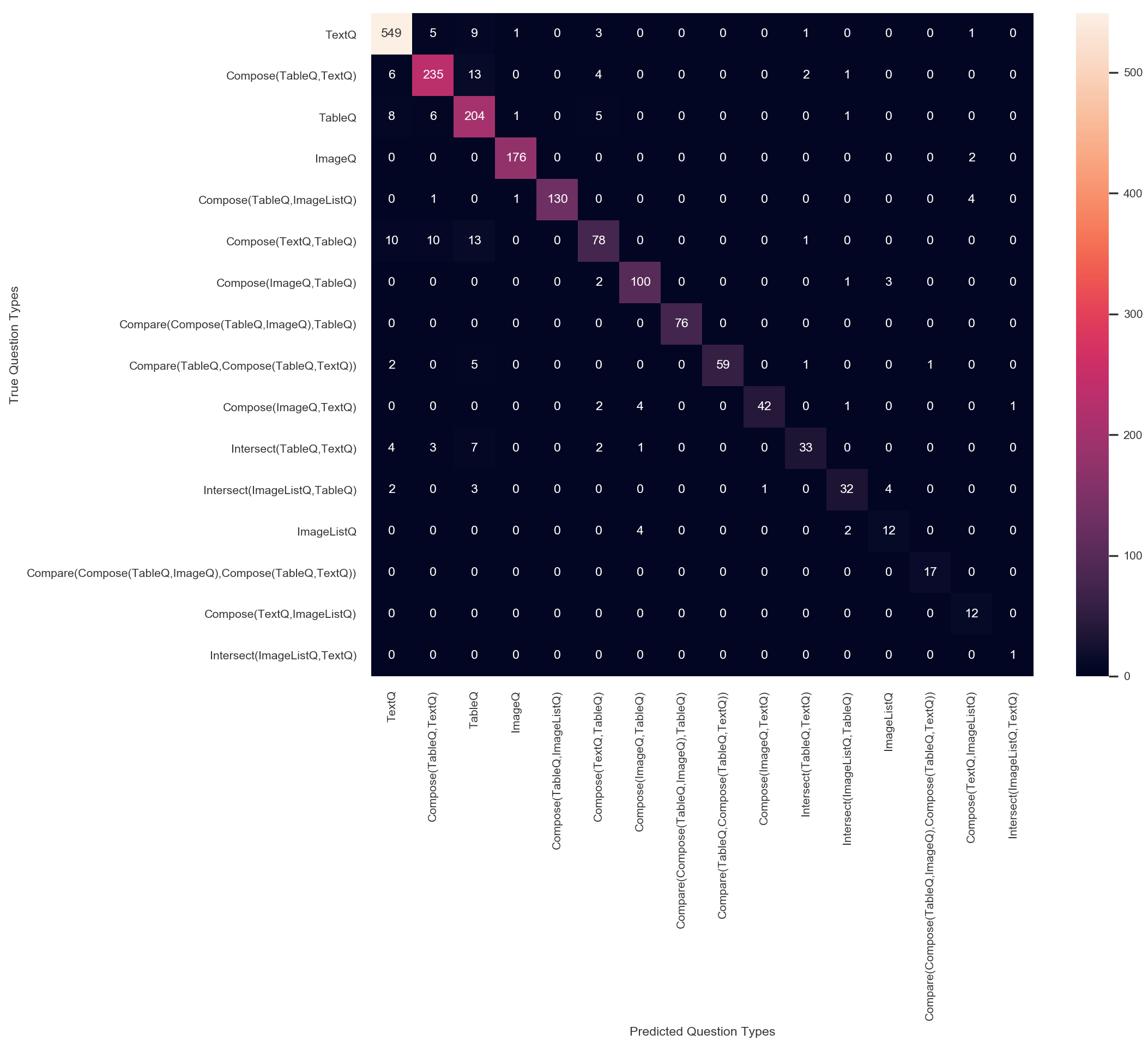}
  \caption{Confusion matrix of the question type classifier on the dev set.}
  ~\label{fig:confusion-fig}
\end{figure}

For our baselines we first train a question-type classifier, based on RoBERTa-large, that takes a question $Q$ as input, and predicts one of the 16 possible question types. Our question type classifier obtains an overall accuracy of 91.4\% on the test set. Figure \ref{fig:confusion-fig} shows the classifier's program prediction confusion matrix, such that predictions on the diagonal are correct, and outside the diagonal are incorrect. We observe several incorrect predictions between the programs containing the \emph{Compose()} operator and TextQ, this is assumed to be due to similar questions in the \emph{HotpotQA} that are used as part of TextQ. 

\begin{figure}[t]
    \begin{center}
  \frame{\includegraphics[width=0.7\columnwidth,height=0.95\textheight]{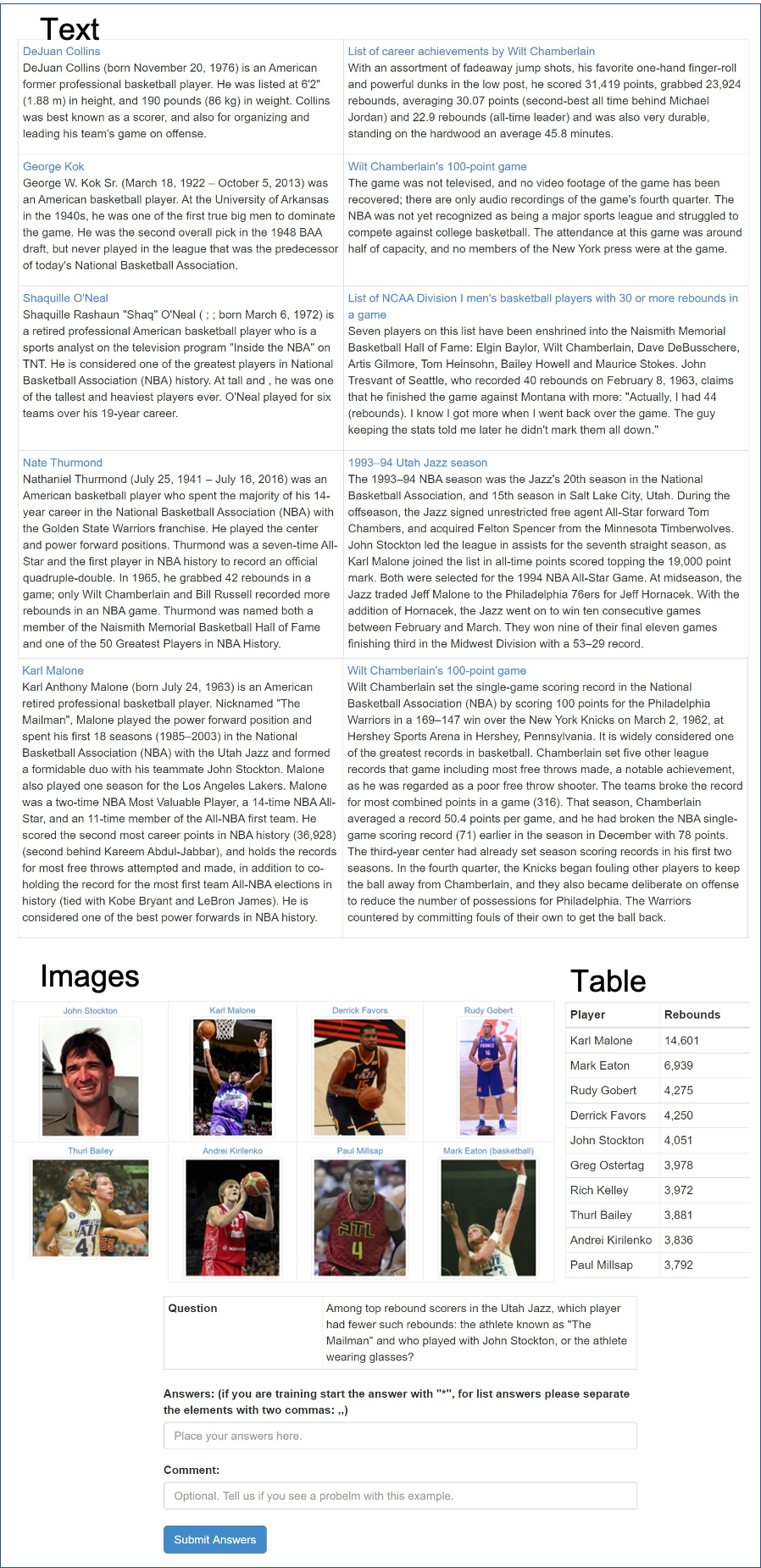}}
  \caption{An example for human evaluation performance task. For each question, we show the relevant context from all the modalities, and ask the user to answer the question.}
  ~\label{fig:amt-ui-human-evaluation}
   \end{center}
\end{figure}

We evaluate the human performance using a special UI that enables the user to move between modalities in the context. Figure \ref{fig:amt-ui-human-evaluation} Illustrates the components in a full context: 10 Text Paragraphs, a list of images and a table. The user may chose on which modality to inspect before providing the final answer. The answer is provided as free text.

%\bibliography{appendix/appendix_iclr2021_conference}
%\bibliographystyle{appendix/appendix_iclr2021_conference}

%\section{Appendix}
%Please see separate file for supplementary material.

\end{document}